\documentclass[10pt,journal,compsoc]{IEEEtran}
\newtheorem{thm}{\bf Theorem}

\newtheorem{game}{\bf Game}
\newtheorem{prop}{\bf Proposition}

\usepackage{amsmath,amssymb,amsfonts}
\usepackage{algorithm}
\usepackage{algpseudocode}

\usepackage{graphicx}
\usepackage{textcomp}
\usepackage{xcolor}
\usepackage{xcolor}
\usepackage{soul}
\usepackage{multirow}
\usepackage{subfigure}
\usepackage[utf8]{inputenc}
\usepackage{bm}
\usepackage{diagbox}

\ifCLASSOPTIONcompsoc
  \usepackage[nocompress]{cite}
\else
  \usepackage{cite}
\fi
\ifCLASSINFOpdf
\else
\fi
\begin{document}
%
\title{Collaboration in Participant-Centric Federated Learning: A Game-Theoretical Perspective}
%
%
%
%

\author{Guangjing Huang, Xu Chen, Tao Ouyang, Qian Ma, Lin Chen and Junshan Zhang
\IEEEcompsocitemizethanks{\IEEEcompsocthanksitem Guangjing Huang, Xu Chen, Tao Ouyang, and Lin Chen are with the School of Computer Science and Engineering, Sun Yat-sen University, Guangzhou 510006, China.
\IEEEcompsocthanksitem Qian Ma is with the School of Intelligent System Engineering, Sun Yat-sen University, Guangzhou 510006, China.
\IEEEcompsocthanksitem Junshan Zhang is with the ECE Department, University of California Davis, United States.}
}
\IEEEtitleabstractindextext{%
\begin{abstract}
Federated learning (FL) is a promising distributed framework for collaborative artificial intelligence model training while protecting user privacy.
A bootstrapping component that has attracted significant research attention is the design of incentive mechanism to stimulate user collaboration in FL. The majority of works adopt a broker-centric approach to help the central operator to attract participants and further obtain a well-trained model.
Few works consider forging participant-centric collaboration among participants to pursue an FL model for their common interests, which induces dramatic differences in incentive mechanism design from the broker-centric FL. To coordinate the selfish and heterogeneous participants, we propose a novel analytic framework for incentivizing effective and efficient collaborations for participant-centric FL.
Specifically, we respectively propose two novel game models for contribution-oblivious FL (COFL) and contribution-aware FL (CAFL), where the latter one implements a minimum contribution threshold mechanism. We further analyze the uniqueness and existence for Nash equilibrium of both COFL and CAFL games and design efficient algorithms to achieve equilibrium solutions.
Extensive performance evaluations show that there exists free-riding phenomenon in COFL, which can be greatly alleviated through the adoption of CAFL model with the optimized minimum threshold.
\end{abstract}

}

\maketitle

\IEEEdisplaynontitleabstractindextext

%
\IEEEpeerreviewmaketitle

\IEEEraisesectionheading{\section{Introduction}\label{sec:introduction}}
\IEEEPARstart{W}{ith} the rapid development of Internet of Things (IoT) and mobile networks,
massive volumes of user data are being generated and geographically scattered over the edges of networks \cite{iotbig2019} \cite{iot2018} \cite{Wireless2019}.
To learn valuable knowledge and information from these massive data, artificial intelligence technology (AI) has been widely used to support machine learning, which dramatically expedites the emergence of many intelligent IoT mobile applications \cite{Marchine2018}.
Traditional machine learning frameworks typically require the uploading local training data to a central server for centralized learning, which suffers from the risk of privacy leaks \cite{privacy2018}. To address such critical issue, federated learning (FL) has been proposed by Google as a novel paradigm to train AI model in a privacy-preserving manner \cite{Com16}, in which the device users learn from their local data and then upload their local model updates to the central server. Since only the local update (e.g., parameter gradients), rather than the local data, is sent to the central server by encrypted communication, the FL enables the device users to preserve data privacy efficiently.

\begin{figure}[t]
	\label{illustration}
	\centering
	\subfigure[Broker-centric Federated Learning]{
		\includegraphics[height=0.2\textwidth]{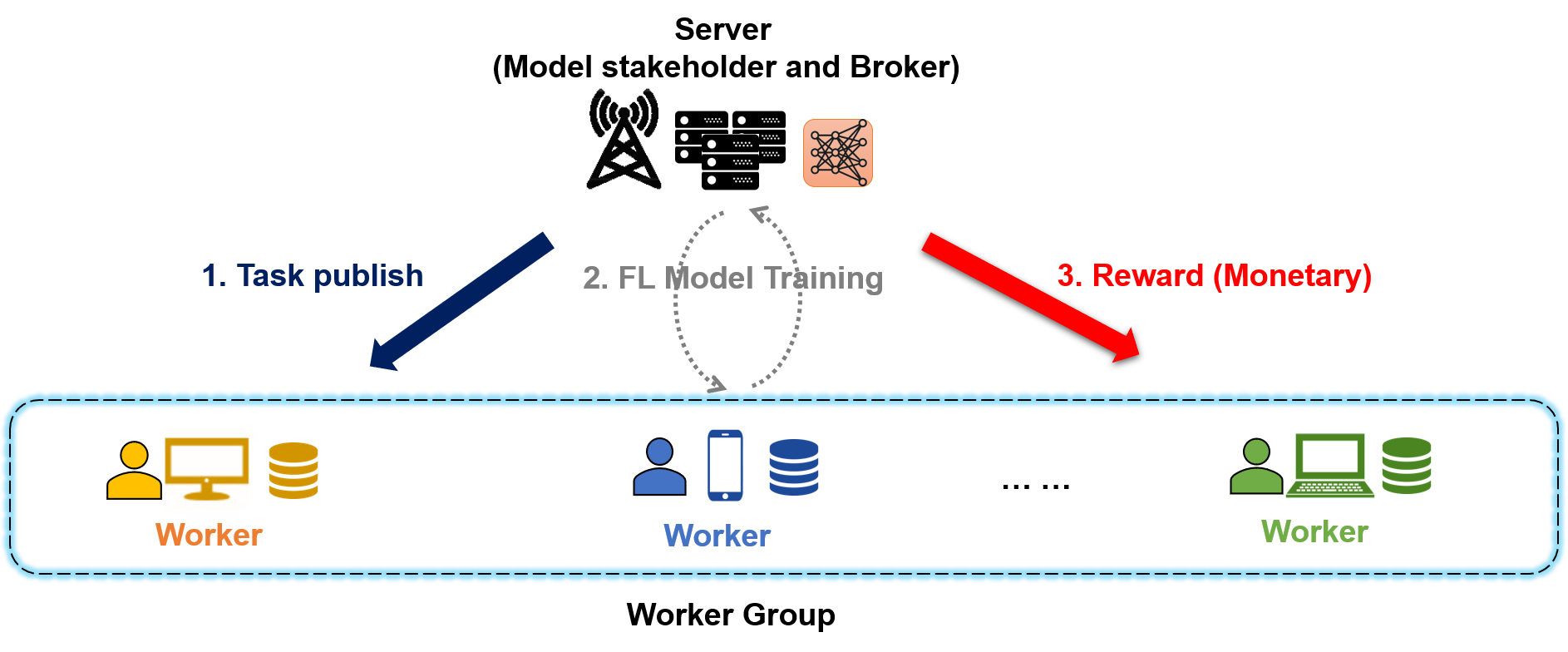}
		\label{workers}
	}
	\subfigure[Participant-centric Federated Learning]{
		\includegraphics[height=0.2\textwidth]{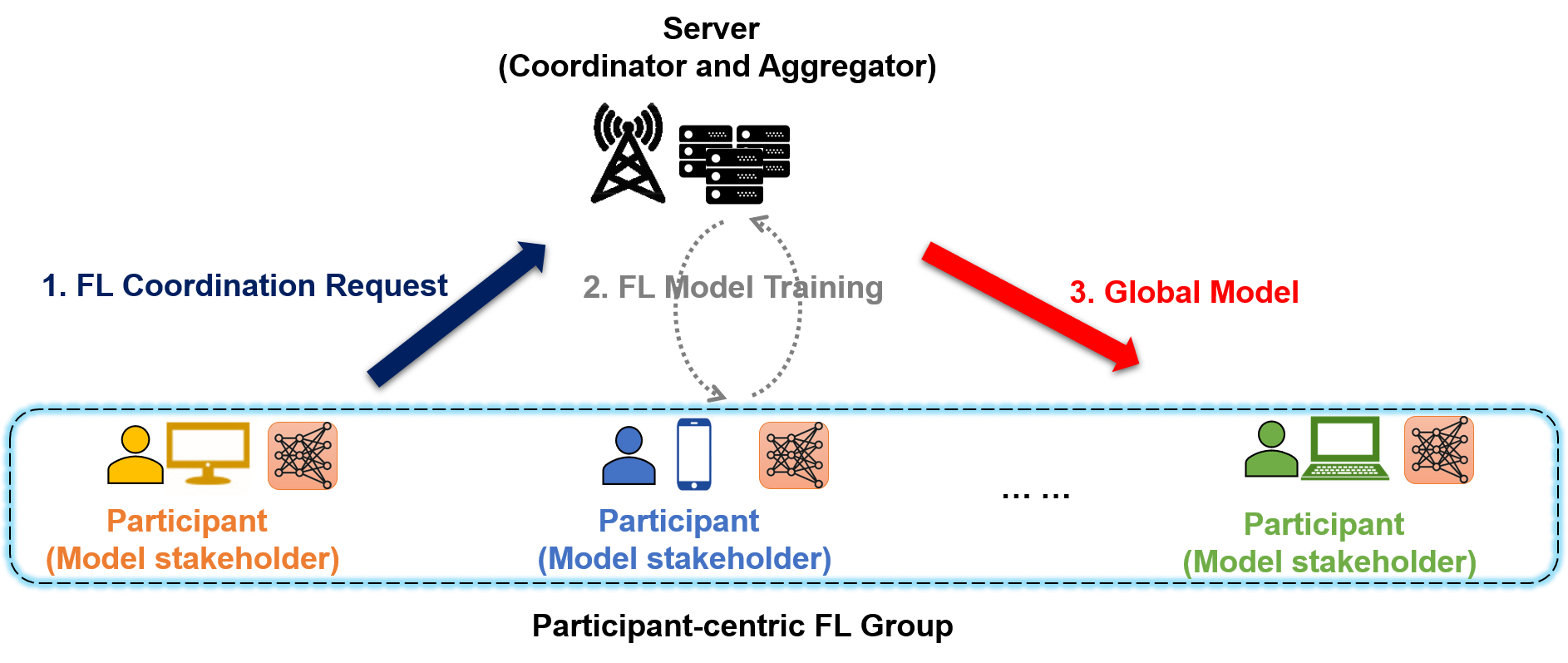}
		\label{participant}
	}
	\caption{An illustration of (a) broker-centric Federated Learning (b) participant-centric Federated Learning.}
	\vspace{-0.2cm}
\end{figure}

Nevertheless, local model training incurs significant costs (e.g., computation cost or energy consumption) for the involved users in FL. Without a proper incentive mechanism, selfish participants would be reluctant to participate in FL. In terms of incentive mechanism design, the majority of existing studies focus on a broker-centric paradigm for incentivizing participation in FL (Fig. \ref{workers}). That is, a central broker publishes a model training task and becomes a model stakeholder (model owner), who attracts participants (workers) to help complete the model training through monetary rewards via economic mechanisms such as auction and contract design
\cite{contract2019, Crow2019, Motivate2019,mofang20,joint2019,DRL2020,Auction21}. In these settings, the participants aim to get more payments from the broker through local model training and uploading.

Along a different line, in many application scenarios, FL training tasks are not generated by a central broker, but initiated by a group of users who themselves are also the participants in the FL \cite{incentive_sur2021,FDML2019}, in order to collectively train a common FL model. In such participant-centric FL (Fig. \ref{participant}), the participants are the model stakeholders and they have the common interest of obtaining a high-quality FL model via mutually-beneficial knowledge sharing. In this case, the FL server works as a coordinator to assist the FL procedure. For example, for smart home applications, users of different families can collectively boost the performance of in-home healthcare AI service via FL \cite{Fedhome20}.

However, how to design a proper incentive mechanism for participant-centric FL is much less understood in the literature \cite{2019advances}. Given that all participants are selfish to act for their own benefits and are also heterogeneous in various perspectives (e.g., valuation on the FL model, data sizes and computing capability), how to obtain an effective strategy for stimulating mutually-beneficial collaboration among the participants is challenging. We should emphasize that, for broker-centric FL, users' heterogeneous cost can be well characterized and compensated with the monetary incentive by the centralized broker via classic two-sided economic mechanisms such as auctions and contracts. However, for participant-centric FL, obtaining a common FL model is the major incentive for the users, but they have usually heterogeneous cost and valuations with complex interaction structures, and their marginal benefits/contributions for each other in the FL model training are very hard to quantify. Hence designing monetary incentive mechanisms that can ensure mutually-satisfactory and trustworthy payment transferring among the users as rewards are generally prohibitive in practice, and thus monetary-free incentive mechanisms with lightweight implementation complexity are much more desired for participant-centric FL. Specifically, we need to address the following key issues for incentivizing collaboration in participant-centric FL.

First of all, a monetary-free incentive-driven collaboration strategy for participant-centric FL is essential, because otherwise the participants acting for their own benefits may deviate from the collaboration strategy.
This requirement motivates a key issue: \textit{How can we model the heterogeneous participants' selfish behaviors and derive an effective collaboration strategy such that each participant has the incentive to follow without external monetary rewards?}

Second, since the trained FL model can be shared among the participants, some participants may benefit from FL but contribute nothing, namely ``free-riding''. The free-riding phenomenon leads to great unfairness in participant-centric FL. This motivates the second key issue:
\textit{How can we design an efficient mechanism to thwart the free-riding behaviors in participant-centric FL?}

Last but not least, given that participant-centric nature of FL in our setting, how to attain a high global efficiency with a low implementation complexity would be a key challenge. This motivates the third key issue: \textit{How can we maximize the total utility of the participants while maintaining the self-stability of FL in a lightweight manner?}

To address the above two key issues, we will leverage the game theoretic modeling approach for developing a comprehensive analytical framework for participant-centric FL. We take into account users' selfish behaviors and heterogeneous characteristics, in order to derive effective collaboration strategies with the desired properties of incentive-driven collaboration, free-riding mitigation and global efficiency boosting for participant-centric FL with a simple minimum contribution threshold mechanism. The contributions of this paper are summarized as follows:
\begin{itemize}
	\item \textit{Contribution-oblivious FL (COFL) game analysis}: To understand participants' selfish behaviors, we first propose a participant-centric FL game called COFL to enable participants  to train a shared federated model collectively without imposing the contributing requirements. We show that the COFL game admits a unique Nash equilibrium under some regularity conditions, and also devise an algorithm to achieve it. However, our findings reveal that the free-riding phenomenon exists in COFL, which would lead to the critical issue of unfairness and greatly harm participants' motivations for collaboration in participant-centric FL.

	\item \textit{Contribution-aware FL (CAFL) game analysis}: To alleviate the free-riding behaviors in participant-centric FL, we then devise a simple and effective collaboration mechanism and propose a novel enhanced game model of CAFL, where a participant will be excluded from getting the trained FL model if his promised training batchsize does not reach the minimum contribution threshold (i.e., a minimum threshold batchsize mechanism). However, we show that the existence of Nash equilibrium cannot be always guaranteed in CAFL, which may cause instable participating behaviors in FL. Thus, we propose an algorithm to refine the set of participants in CAFL, which can always guarantee to achieve a Nash equilibrium of the refined CAFL game. Furthermore, we boost the global performance by finding the optimal threshold to maximize the total utility of all participants.

	\item \textit{Extensive performance evaluation}: We finally conduct extensive numerical evaluations with the realistic MNIST and CIFAR10 datasets to verify our theoretical analysis and results. We find that, compared with COFL, CAFL not only effectively alleviates the free-riding phenomenon, but also significantly boost the amount of participation and total utility to a great extent, stimulating more than 90\% participants to contribute with a superior performance very closing to the solution of optimal total utilities in most cases.
\end{itemize}

The rest of this paper is organized as follows. Section 2 introduces FL framework and problem formulation respectively. Contribution-oblivious FL (COFL) is presented in Section 3. We discuss Contribution-aware FL system and algorithm design in Section 4 and 5 respectively. Numerical result is showned in Section 6. We introduce related work in Section 7 followed by conclusion in Section 8.

\section{System Model}
We consider the participant-centric FL formed by multiple device users at the network edge. We adopt the standard federated learning framework in \cite{Com16, Acc19} and define a global iteration as follows.
During the local training phase, all device participants calculate local gradients based on the their data and the received model. Then the edge server (i.e., central server) aggregates all local gradients from participants (i.e., the device users) and broadcasts new gradients to each participant for next local model updates.  This complete update process is called a global iteration.
All participants repeat the global iteration until the global model converges.
Since model training incurs significant time and energy cost, participants should carefully strike a balance between their valuation on the FL model (i.e., getting the model as rewards) and the training costs.
In what follows, we first characterize participants' training costs, and then calculate participants' utilities.

\subsection{Participant Cost Model}
In a global iteration, a participant's model training cost consists of computation and communication cost.

\textbf{Computation cost}: The set of $K$ participants is denoted as $\mathcal{K}= \left\lbrace 1,2,...K\right\rbrace $. Each participant $k$ owns a local dataset $\mathcal{D}_{k}=\{ (\boldsymbol{x}_{k}^{1},y_{k}^{1}),...,(\boldsymbol{x}_{k}^{N_{k}},y_{k}^{N_{k}})\}$, and $|\mathcal{D}_{k}|$ denotes the data size of participant $k$, and $\boldsymbol{x}^{i}_{k}$ and $y_{k}^{i}$ represent the model features and the Ground truth label of participant $k$'s $i$-th data respectively.
Similar to most existing approaches such as \cite{Acc19}, we adopt the mini-batch stochastic gradient descent (SGD) algorithm for local training. Each participant selects a subset of the local dataset, called one batch, to calculate local gradients in one global iteration.
The number of data samples in one batch selected by participant $k$ is $B_{k}$ and the global batchsize is $B=\sum_{k\in\mathcal{K}}B_{k}$. Obviously, $0\leq B_{k}\leq B_{k}^{max} \overset{\Delta}{=} |\mathcal{D}_{k}|$.
To model a participant's energy consumption, we further define $C_{k}$ as the average number of CPU cycles for participant $k$ performing local gradient calculation with one data sample.
Moreover, $f_{k}$ is the adopted CPU processing speed (i.e., computing resource) of participant $k$ for local model training. In general, we have $f_k^{min}\leq f_k \leq f_k^{max}$ where $f_k^{min}$ and $f_k^{max}$ represent participant $k$'s minimum and maximum computing capabilities, respectively.
Based on the above definitions, the energy consumption of local gradients calculation of participant $k$ in one global iteration can be calculated as \cite{Fed19}:
\begin{equation}\label{Ecom}
E_{k}^{com}=\dfrac{1}{2} \alpha C_{k} B_{k} f_{k}^{2},
\end{equation}
where $\alpha$ is a coefficent corresponding to the computing chip architecture. Also, the local training process latency can be calculated as:
\begin{equation}\label{Tcom}
T_{k}^{com}=\dfrac{C_{k} B_{k} }{f_{k}}.
\end{equation}

\textbf{Communication cost}: At the phase of data communication,
the local gradients are sent to the center server through the multiple channel access technology (e.g., orthogonal frequency division medium access (OFDMA)) once the local model training accomplished.
The communication resource allocation optimization is usually determined by the central server instead of participants. Since we focus on analyzing the participants' behaviors in participant-centric FL,  we assume that the communication resource allocation for the participants' data transmission is fixed.
We also assume that the data size of the model parameter gradients is the same for each participant in each FL iteration (which is usually the case since all the participants train the same FL model).
Thus, in this study we hence assume the communication cost is fixed for each participant in the following discussions \footnote{The communication cost for each participant can be different and this case can be captured by adding a participant-specific constant $D_k$ into the participant's utility function in \eqref{UP}.  Nevertheless, since the communication cost $D_k$ is fixed and does not impact participant's decision on the computation resource allocation and batch size selection, we will neglect the fixed communication cost in the following analysis for simplicity. This simplicity is common in many exiting studies related to incentive mechanism design \cite{Contract21} \cite{Robust2022} \cite{Trade20} and conducive to directly revealing economic characteristic for mechanism design, since the communication part is independent of FL model performance.
}.

\subsection{Participant Utility Function}
We next turn attention to the participants' utilities in participant-centric FL.
In general, all participants aim to obtain a well-trained model with a low loss function (or high model accuracy).
In one global iteration, higher model accuracy requires a larger global batchsize, which will further induce a higher training cost (i.e., local energy consumption and latency).
In this regard, according to \cite{Acc19}, we first define the global loss function and the global loss decrease (model accuracy improvement) as follows:
\begin{equation}\label{L}
L\left( \boldsymbol{w} \right)=
\dfrac{1}{|\bigcup_{k\in \mathcal{K}}\mathcal{D}_{k}|}
\sum_{k\in\mathcal{K}}|\mathcal{D}_{k}|
L_{k}\left( \boldsymbol{w}, \mathcal{D}_{k} \right),
\end{equation}
\begin{equation}\label{triL}
\triangle L [n]=L( \boldsymbol{w}\left[n\right])-
L(\boldsymbol{w}\left[n-1\right]),
\end{equation}
where $L_{k}\left( \boldsymbol{w}, \mathcal{D}_{k} \right) $ is a general convex loss function of participant $k$, wherein $\boldsymbol{w}$ is the machine learning model parameter. $L\left( \boldsymbol{w}\left[n\right] \right)$ is the loss function of $n$-th global iteration. We consider one global iteration and replace $\triangle L[n]$ with $\triangle L$. We can measure the model accuracy improvement in one global iteration by $\triangle L$ which depends on global batchsize. According to \cite{BL14}, the expected difference $f(\boldsymbol{w}[n])-f(\boldsymbol{w}^*)$ is bounded by $O(1/\sqrt{Bn}+1/n)$, when participants use mini-batch SGD in the IID case, where $B$ is global batchsize and $n$ is the number of global iteration. Given fixed $n$, the upper bound of expected difference is decreasing and convex function respect to global batchsize $B$ and satisfies diminishing marginal effect. In one global iteration, the model improvement $\triangle L$ is dominated by term $\sqrt{B}$ and can be approximately expressed as \cite{Acc19}:
\begin{equation}\label{Lbatch}
\triangle L=\xi\sqrt{B}=\xi\sqrt{\sum_{k\in\mathcal{K}}B_{k}},
\end{equation}
where $\xi$ depends on the model structure. For simplicity, we use $\xi=1$ in this paper because it is a scaling constant for a given training model. Intuitively, a large amount of training data leads to a good model performance in one global iteration.
Based the above definitions, the utility function for participant $k$ is
\begin{equation}\label{UP}
U_{k} \left( B_{k} , f_{k}\right)=\theta_{k} \ln(1+\triangle L)-\varphi_{k} E_{k}^{com}-\gamma_{k} T_{k}^{com}.
\end{equation}
Here, $\theta_{k}$ describes participant's preference of the FL model (i.e., a larger $\theta_k$ implies that  participant $k$ has a higher valuation on the model). $\varphi_{k}$ and $\gamma_{k}$ are weight parameters corresponding to participant $k$'s energy consumption and latency of local training processing respectively. Note that in \eqref{UP} we omit the decision variables of other participants for notational simplicity.
The utility function \eqref{UP} captures that a participant should jointly consider the model improvement, energy consumption and local training time. Corresponding to the economic characteristics of the upper bound of expected difference, the term $\theta_{k} \ln(1+\triangle L)$ is increasing and concave function respect to global batchsize $B$ and reveals the rule of diminishing marginal returns for model improvement \cite{Incent12}. The
physical meaning of the logarithmic function is that the magnitude of FL model’s improvement decreases with global
batchsize. That is, when a FL model possesses a large global batchsize, participants make the same effort but obtain
a low return, , which would encourage participants to contribute more when the global batchsize is small.
Note that, for ease of exposition, we adopt a specific logarithmic function in this paper.
Actually, our theoretical results and analysis still hold for a general function $\theta_{k} g(B)$ wherein the model accuracy related function $g(B)$ is a second-order differentiable function and satisfies diminishing marginal effect ($g'(B)>0$, $g''(B)<0$ and $g'(0)>\frac{A_{k}}{\theta_{k}}$). Here, $A_{k}$ is the unit training cost (which will be discussed later). The discussion on general function is given in Subsection A in Appendix C in the separate supplementary file.

Assuming the FL process converges in a finite number of global iterations\footnote{The number of global iterations for a fixed model accuracy depends on global batchsize. In general, the global batchsize is much larger than the local batchsize of a participant, Thus, we assume that a single participant's decision has little impact on the number of global iterations.}, we define the utility function of one global iteration in \eqref{UP} to estimate a participant's utility throughout the training process.
Based on \eqref{Ecom}\eqref{Tcom}\eqref{Lbatch}\eqref{UP}, each participant $k$ aims to maximize his utility function by tuning their local training batchsize and CPU processing speed, which can be calculated as:
\begin{eqnarray}\label{user}
\notag
U_{k} \left( B_{k} , f_{k} \right)&=&\theta_{k} \ln(1+\sqrt{B_{-k}+B_{k}})-\dfrac{1}{2} \varphi_{k} \alpha C_{k} B_{k} f_{k}^{2}\\
&\,&-\gamma_{k} \dfrac{C_{k} B_{k} }{f_{k}},
\end{eqnarray}
where $B_{-k}=\sum_{i\in\mathcal{K} \setminus\left\lbrace k\right\rbrace} B_{i}$ shows that a participant utility is also influenced by other participants' decisions in terms of the choice of batchsize for local training. For ease of presentation, we predefine $A_{k}=\dfrac{1}{2} \varphi_{k} \alpha C_{k} f_{k}^{\ast2}+ \gamma_{k} \dfrac{C_{k}}{f_{k}^{\ast}} $ as the unit training cost of participant $k$, where $f_{k}^{\ast}$ is the optimal CPU frequency (which will be discussed later).

Note that in this paper, we assume that all participants are selfish but have no intention of sabotaging the FL model. The security issue of FL model is not the focus of this paper. We also assume that all participants’ data are independent and identically distributed (IID). The scenario of non-IID data distributions will be considered in a future work.

In the following, aiming at modeling participants' strategic behaviors in the participant-centric FL, we leverage the game theoretical approach to derive useful insights and devise efficient collaboration strategies accordingly.

\section{Contribution-oblivious FL GAME}
We first consider the contribution-oblivious FL (COFL) game model, in which all participants act on their own benefits to decide the computing resources and data batchsize for local training without imposing any contributing requirements.
Due to the heterogeneous nature among participants (e.g., differences in model preference and training cost), forging an incentive-driven collaboration strategy such that all participants are mutually satisfied is non-trivial. We will analyze the Nash equilibirum of COFL game, propose an algorithm to achieve Nash equilibrium in COFL game, and derive some insightful results.

\subsection{Game Formulation and Best Response in COFL}
We formally define the COFL game as follows:
\begin{game}[COFL Game]\quad
	\begin{itemize}
		\item {\verb|Players|}: The set $\mathcal{K}$ of participants.
		\item {\verb|Strategies|}: The chosen batchsize $B_{k}\in [0, B_{k}^{max}]$, and computing speed $f_{k}\in [f_{k}^{min}, f_{k}^{max}]$ for each $k\in\mathcal{K}$.
		\item {\verb|Utilities|}: The utility $U_k(B_{k},f_{k})$ for each $k\in \mathcal{K}$.
	\end{itemize}
\end{game}

In the following, we combine the choices of data batchsize $B_k$ and computing speed $f_k$ as a computation strategy $\sigma_{k}(B_{k},f_{k})\in \bm{\Sigma_{k}}$, where $\bm{\Sigma_{k}}$ is the participant $k$'s strategy space. Here, we treat $B_k$ as real number, the strategy space is thus a convex hull.

The COFL game reaches Nash equilibrium if and only if none of participants can unilaterally change the strategy to improve his utility. A Nash equilibrium solution is a strategy profile $(\sigma_{1}^{\ast},...,\sigma_K^{\ast})$ such that $\forall k \in \mathcal{K}$, the following inequality holds:
\begin{eqnarray}
U_{k}(\sigma_{1}^{\ast},...,\sigma_{k},...,\sigma_K^{\ast}) \leq U_{k}(\sigma_{1}^{\ast},...,\sigma_{k}^{\ast},...,\sigma_K^{\ast}), \forall \sigma_{k} \in \bm{\Sigma_{k}}. \!\!
\end{eqnarray}
This implies all participants take the mutually best response strategy simultaneously in the FL model training, i.e.,

\begin{eqnarray}
\hat{\sigma}^{\ast}_{k}=\mathop{\arg\max}\limits_{\sigma_{k} \in \bm{\Sigma_{k}}}U_k(\sigma^{\ast}_{1},...,\sigma_{k},...,\sigma^{\ast}_K).
\end{eqnarray}
For the ease of practical implementation, we only consider the pure Nash equilibrium in this paper\footnote{We adopt the widely used Nash equilibrium as the result of game analysis. The inefficient cooperation among the participants in Nash equilibrium is the motivation for our subsequent improvements.}.
To characterize the existence of Nash equilibrium, we first derive participant's the best response strategy in COFL.
Given others' decisions, the best response function of participant $k$ is shown as below.

\begin{prop}
	\textit{The best response strategy $\sigma_{k}$ of participant $k$ in COFL is $(B_{k}^{\ast},f_{k}^{\ast}$):
		\begin{eqnarray}\label{fopt}
		f_{k}^{\ast}=\left[  \sqrt[3]{\dfrac{\gamma_{k}}{\varphi_{k}\alpha} }\right]_{f_{k}^{min}}^{f_{k}^{max}},
		\end{eqnarray}
		and
		\begin{eqnarray}\label{bestre}
		B_{k}^{\ast}=
		\left[  \tilde{B_{k}}\right]_{0}^{B_{k}^{max}},
		\end{eqnarray}
		where $[X]_{Y}^{Z}=\min\left\lbrace \max\left\lbrace Y,X\right\rbrace ,Z\right\rbrace$,
		 $A_{k}=\dfrac{1}{2} \varphi_{k} \alpha C_{k} f_{k}^{\ast2}+ \gamma_{k} \dfrac{C_{k}}{f_{k}^{\ast}} $,
		and
		\begin{equation}\label{Bpai}
		\tilde{B_{k}}=\left(\sqrt{\frac{1}{4}+\frac{1}{2}\frac{\theta_{k}}{A_{k}}}-\frac{1}{2}\right)^2-B_{-k}.
		\end{equation}
	}
\end{prop}

The proof is given in Subsection A in Appendix A in the separate supplementary file.
Since it can be seen from \eqref{fopt} that a participant's optimal CPU frequency is independent of others' strategies, the key issue of the COFL game is how coordinate participants' data batchsizes to achieve a Nash equilibrium.

\subsection{Nash Equilibrium in COFL}
In this subsection, we focus on finding the Nash equilibrium of COFL game. Specifically, according to best response functions in Proposition 1 above, we first capture the characteristic of the Nash equilibrium of the COFL game (Theorem \ref{thm1}), and based on which, we then obtain equivalent form of Nash equilibrium (Theorem \ref{thm2}). We next show the existence and uniqueness of the Nash equilibrium of the COFL game (Theorem \ref{thm3}). Finally, the equilibrium finding algorithm is given in Algorithm \ref{open}.

For simplicity, we rewrite \eqref{Bpai} as $\tilde{B_{k}}=\beta_{k}-B_{-k}$ where
\begin{equation}\label{beta}
\beta_{k} =\left(\sqrt{\frac{1}{4}+\frac{1}{2}\frac{\theta_{k}}{A_{k}}}-\frac{1}{2}\right)^2
\end{equation}
Intuitively, $\beta_{k}$ is a parameter indicating  participant's quality, which captures the mapping relationship from model preference and training cost to participant's batchsize strategy, e.g., a larger $\beta_{k}$ means a higher participating enthusiasm (larger $\tilde{B_{k}}$).

Since participants' model preferences $\theta_k$ are continuous variables in general, the probability that two participants have exactly the same preference would be zero under a given distribution. Hence, we focus on the case  that the set of participants $\mathcal{K}= \left\lbrace 1,2,...,K\right\rbrace $ are sorted in descending order by $\beta_{k}$ without ties, i.e., $\beta_{1}>\cdots>\beta_{K}$.
Best response functions in \eqref{bestre} imply that  there are three types of participants when the game reaches equilibrium. Correspondingly, we define the \textit{type-1} ($S_{1}$), \textit{type-2} ($S_{2}$) and \textit{type-3}  ($S_{3}$) participants. Specifically, $S_1=\{ i\in{\mathcal K}  | B_i^{'}> B_i^{max}\} $, $S_2=\{ i\in{\mathcal K}  | 0 \leq B_i^{'}\leq B_{i}^{max} \}$ and $S_3=\{ i\in{\mathcal K}  | B_i^{'}< 0 \}$.
We can show Theorem \ref{thm1}, which reveals the structural properties of Nash equilibrium in COFL.
\begin{thm}\label{thm1}
	\textit{At the Nash equilibrium of the COFL game, the participants can be divided into the sets of type-1, type-2 and type-3.
		For any $i\in S_{1}, c\in S_{2}, j\in S_{3}$, we have $\beta_{j}<\beta_{c}<\beta_{i}$. Moreover, at most one participant belongs to type-2 participant.}
\end{thm}

\begin{figure}[t]
	\setlength{\abovecaptionskip}{-0.05cm}
	\centering
	\includegraphics[height=0.3\linewidth]{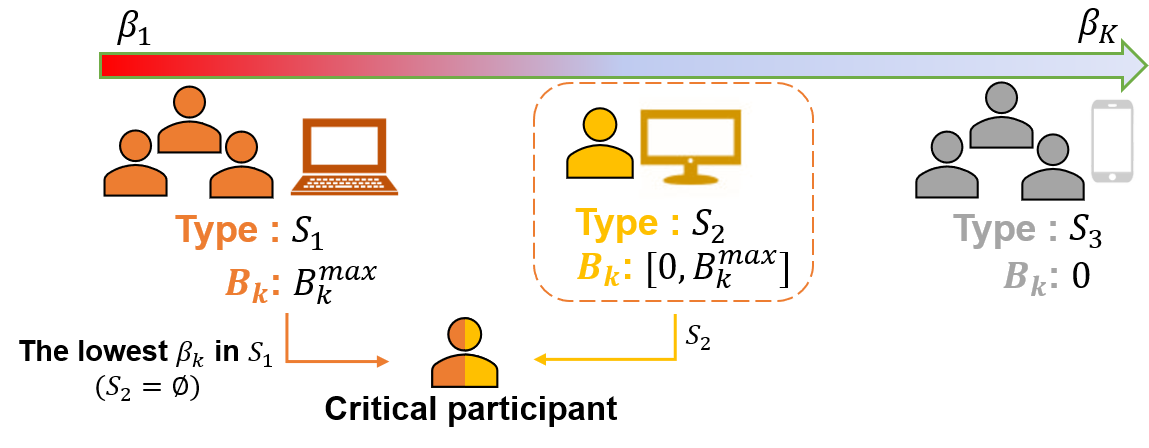}
	\caption{The structure properties of Nash equilibrium in the COFL game.}
	\label{COFLNE}
\end{figure}

The proof is given in Subsection B in Appendix A in the separate supplementary file. As illustrated in Fig. \ref{COFLNE}, Theorem \ref{thm1} indicates when the game reaches the Nash equilibrium (if it exists), $S_{1}$, $S_{2}$ and $S_{3}$ are arranged in descending order of $\beta_{k}$. Among them, the type-2 participant is no more than one. As a result, we attempt to find a boundary between $S_{1}$ and $S_{3}$ called critical participant:

\textbf{Definition 1}: Participant $c$ is a \textbf{critical participant} in COFL if and only if $\forall i\in \mathcal{K}, \beta_{i}<\beta_{c}$, $\tilde{B_{i}}=\beta_{i}-B_{-i}< 0$ and  $\tilde{B_{c}}=\beta_{c}-B_{-c}\geq 0$ when the game reaches a Nash equilibrium (if it exists).

The critical participant belongs to type-1 and has the lowest $\beta_{k}$ among type-1 participants if $S_{2}$ is a empty set. On the other hand, if participant $c$ belongs to type-2, he must be the critical participant. That is, the critical participant is the boundary between $S_{1}$ and $S_{3}$. At Nash equilibrium, participants with $\beta_{k}$ higher than $\beta_{c}$ belong to $S_{1}$. Otherwise, participants belong to $S_{3}$.
Based on Theorem \ref{thm1}, the critical participant has a vital role in the Nash equilibrium, since others must belong to $S_{1}$ or $S_{3}$. For the convenience of
presentation, we predefine of participant $c$'s equilibrium structure in COFL game.

\noindent\textbf{Definition 2 (participant's equilibrium structure)}: Given a participant $c \in \mathcal{K}$, (1) $\forall i \in \mathcal{K}$, $\beta_{i}>\beta_{c}$, we set $B_{i}=B_{i}^{max}$. (2) $\forall j \in \mathcal{K}$, $\beta_{j}<\beta_{c}$, we set $B_{j}=0$. We call the strategy profile ($B_{1},...,B_{c-1},B_{c+1},...,B_{K}$) as the participant $c$'s equilibrium structure, where each $B_{i} (i\neq c)$ satisfy (1) and (2).

We consider the conditions under which a participant $c$ becomes the critical participant at equilibrium and derive the equivalent form of Nash equilibrium in Theorem \ref{thm2}.

\begin{thm}\label{thm2}
	If there exists a participant c with equilibrium structure ($B_{1},...,B_{c-1},B_{c+1},...,B_{K}$), then one of the following properties holds for the COFL game:
	\begin{itemize}
		\item {\verb|Property 1|}: $B_{c}^{\ast}=B_{c}^{max}$ and $\beta_{c+1} <B_{nash} <\beta_{c}$.
		\item {\verb|Property 2|}: $B_{c}^{\ast}=\tilde{B_{c}}$ and $B_{nash}=\beta_{c}$.
	\end{itemize}

	Where $B_{nash}$ is the global batchsize (i.e., total batchsize contributed by the participants) at the equilibrium. In this case, Nash equilibrium exists in the game, and the participant $c$ is critical participant. Also, if the COFL game reaches a Nash equilibrium (if it exists), one of the two properties is satisfied.
\end{thm}

The proof is given in Subsection C in Appendix A in the separate supplementary file.
In Theorem \ref{thm2}, Property 1 considers a case where critical participant $c$ belongs to \textit{type-1} and Property 2 corresponds to $c$ as a \textit{type-2} participant. Based on Theorem \ref{thm2}, we can search an equilibrium through considering the relationships between global batchsize and $\beta_{k}$.
To account the global batchsize by all the participants, we define a global batchsize function $F_o(c,B_{c}^{con})$ in terms of the critical participant $c$ and its batchsize $B_c^{con}$ as:
\begin{align}
F_o(c,&B_{c}^{con})=\sum_{k=1}^{c-1}B_{k}^{max}+B_{c}^{con}+\sum_{k=c+1}^{K}0 ,\label{Fopen}
\\ &\mbox{s.t.}\quad c \in \mathcal{K},  \tag{\ref{Fopen}{a}} \label{Fopena}
\\ &  \quad \quad \, 0\leq B_{c}^{con}\leq B_{c}^{max}. \tag{\ref{Fopen}{b}} \label{Fopenb}
\end{align}
The first and third terms in the function \eqref{Fopen} originates the batchsize contributions by \textit{type-1} and \textit{type-3} participants according to Theorem \ref{thm2}. For the feasibility of such algorithm design, we explore the existence and uniqueness of Nash equilibrium, which is guaranteed by Theorem \ref{thm3}.

\begin{thm}\label{thm3}
	\textit{The contribution-oblivious FL game must admit a unique Nash equilibrium.}
\end{thm}

The proof is given in Subsection D in Appendix A in the separate supplementary file.
Based on above analysis, we introduce our approach on how to compute the Nash equilibrium in COFL in Algorithm 1. The key idea is to search a participant $c$ satisfying Property 1 or 2 in Theorem \ref{thm2}. If we have found the critical index $c$ and his strategy in Nash equilibrium, we can then know others' optimal strategies, i.e., $\beta_{i}>\beta_{c},B_{i}=B_{i}^{max}$ and $\beta_{j}<\beta_{c},B_{j}=0$.
As shown in Algorithm \ref{open}, we search the lowest index $i$ satisfying $F_o(i,B_{i}^{max})>\beta_{i}$. Here, $i-1$ or $i$ is the critical participant. Line 9 and 10 in Algorithm \ref{open} calculate the critical participant's strategy when others' are fixed.

\begin{algorithm}[t]
	\caption{COFL Nash Equilibrium Computing Algorithm}
	\label{open}
	\begin{algorithmic}[1]
		\Require $K$ participants (descending order by $\beta_{k}$)
		\Ensure Strategy profile $(B_{1},B_{2},...,B_{K})$
		\State Search lowest index $i$ satisfying $F_o(i,B_{i}^{max})>\beta_{i}$ using binary search.
		
		\If{$F_o(i,0) \leq \beta_{i}$}
		\State $i$ is the critical participant denoted as $c$.
		\Else
		\State $i-1$ is the critical participant denoted as $c$.
		\EndIf
		
		\State $\forall k\in \mathcal{K}, \beta_{k}>\beta_{c}$, we set $B_{k}=B_{k}^{max}$.
		\State $\forall k\in \mathcal{K}, \beta_{k}<\beta_{c}$, we set $B_{k}=0$.
		
		\State Calculate $\tilde{B_{c}}$ using \eqref{Bpai}.
		\State Calculate $B_{c}$ using \eqref{bestre}.
		\State \textbf{return} strategy profile $(B_{1},B_{2},...,B_{K})$
	\end{algorithmic}
\end{algorithm}

\subsection{Free-Riding Phenomenon}

\begin{figure}[t]
	\centering
	\includegraphics[height=0.25\textwidth]{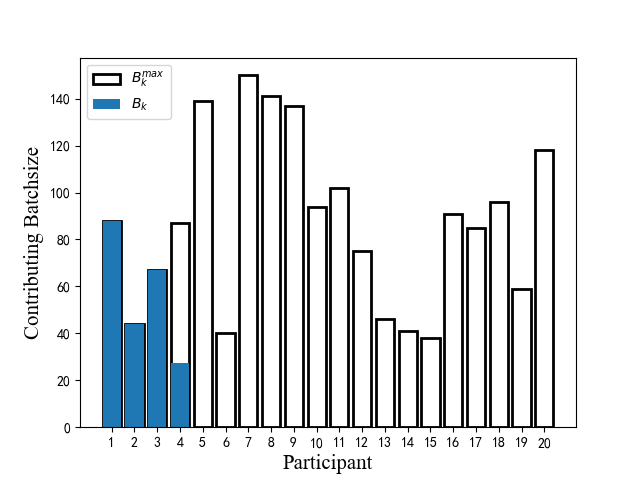}
	\caption{The ``free-riding'' phenomenon in COFL.}
	\label{OAstra}
\end{figure}

When COFL reaches the Nash equilibrium, Theorem \ref{thm1} shows that \textit{type-1} participants contribute their maximum data batchsize in model training. However, the \textit{type-3} participants can obtain the FL model without any cost, which raises the critical issue of free-riding.
By executing Algorithm \ref{open}, Fig. \ref{OAstra} shows the free-riding phenomenon of a group, which is made up of 10 \textit{High quality} and 10 \textit{Low quality} participants (the setting of these two kinds of participant is detailed in Section 6). The critical participant is 4. As a result, the participants with $\beta$ less than  $\beta_{4}$ (participants indexing from 5 to 20) choose to free-ride and contribute nothing. This illustrates that the free-riding problem by a significant portion of participants would greatly break the fairness and harm the motivations for collaboration in participant-centric FL if we directly apply COFL game model based on collaboration strategy.

\section{Contribution-aware FL GAME}
\begin{figure}[t]
	\centering
	\includegraphics[height=0.2\textwidth]{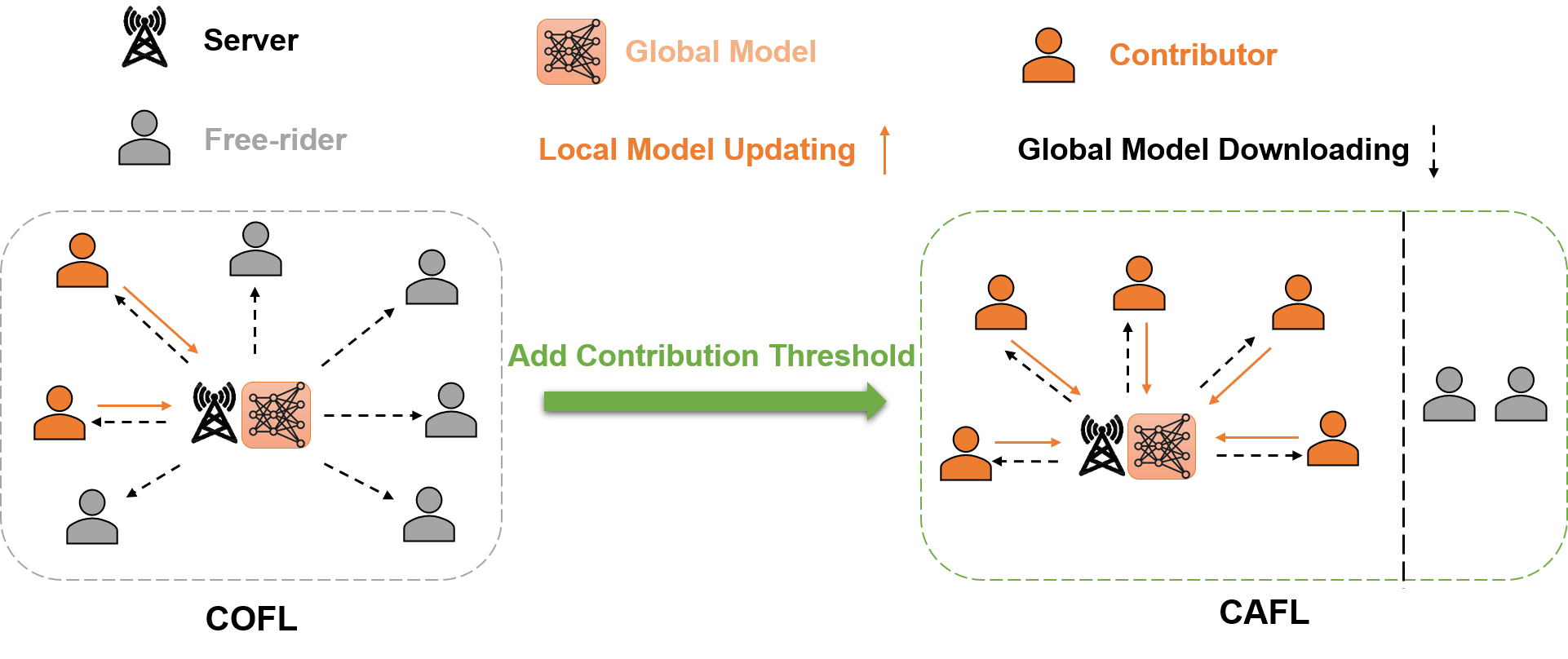}
	\caption{The comparison of COFL and CAFL.}
	\label{d_coafls}
\end{figure}

In response to the challenges of free-riding phenomenon in COFL, we then propose a novel contribution-aware FL (CAFL) game model, where a participant will be excluded from FL if his contributed batchsize is lower than a given minimum threshold batchsize $B_{th}$. As illustrated in Fig. \ref{d_coafls}, when a participant is excluded, he cannot receive the trained FL model from the server. As a result, more participants are willing to contribute in order to obtain the FL model as the return.
In the followings, we will formulate the CAFL game and focus on investigating the equilibrium solution.

\subsection{Game Formulation and Best Response in CAFL}

Different from COFL, we add a minimum threshold mechanism in order to mitigate the free-riding issue by imposing the minimum contribution requirement.
The minimum threshold satisfies $0<B_{th}\leq B_{th}^{max}= \underset{\forall k \in \mathcal{K}}{\min}(B_{k}^{max})$. Given a $B_{th}$, the utility function of participant $k$ is defined as:
\begin{eqnarray}\label{Cuser}
U_{k} \left( B_{k} , f_{k} \right)=\left\{
\begin{aligned}
&\theta_{k} \ln(1+\sqrt{B_{-k}+B_{k}})-\dfrac{1}{2} \varphi_{k} \alpha C_{k} B_{k} f_{k}^{2}\\
&-\gamma_{k} \dfrac{C_{k} B_{k} }{f_{k}}, \quad \mbox{if} \  B_{k}\in \left[B_{th}, B_{k}^{max} \right]. \\
&0, \quad \mbox{if} \ B_{k}=0.
\end{aligned}
\right.
\end{eqnarray}
Here, $U_{k}\left(0, f_{k}\right)=0$ means that a participant contributes nothing and is excluded to participate in FL with zero utility. The discussion of strategy in interval (0, $B_{th}$) is meaningless, since the participant will be excluded from FL. We then define the CAFL game model as follows:

\begin{game}[CAFL Game]\quad
	\begin{itemize}
		\item \it{Players}: The set $\mathcal{K}$ of participants.
		\item \it{Strategies}: $B_{k}\in \left\lbrace 0 \right\rbrace \cup [B_{th}, B_{k}^{max}] $, $f_{k}\in [f_{k}^{min}, f_{k}^{max}]$.
		\item \it{Utilities}: The utility $U_k(B_{k},f_{k})$ in \eqref{Cuser} for each $k\in \mathcal{K}$.
	\end{itemize}
\end{game}

In CAFL, we assume that if participant $k$ participates in CAFL and his utility $U_{k} \geq 0$, he is willing to participate in the FL. Accordingly, we derive the conditions for participants to participate in FL as follows:

\begin{prop}\label{pro2}
	\textit{A participant $k$ is willing to  participate in federated learning in CAFL if and only if the following holds:}
	\begin{eqnarray}\label{Botherk}
	B_{-k}\geq ( e^{\frac{A_{k}B_{th}}{\theta_{k}}} -1)^{2}-B_{th}.
	\end{eqnarray}
\end{prop}
The proof is given in Subsection A in Appendix B in the separate supplementary file.
For simplicity, we define a function $\Phi$ to represent the right hand of \eqref{Botherk} as:
\begin{eqnarray}
\Phi ( B_{th},  \frac{\theta_{k}}{A_{k}} )=(e^{\frac{A_{k}B_{th}}{\theta_{k}}} -1)^{2}-B_{th}\triangleq \Phi_{k}.
\end{eqnarray}

Proposition \ref{pro2} shows that whether a participant participates in FL depends on $B_{th}$, participant's parameter $\theta_{k}$ and unit training cost $A_{k}$. From \eqref{Botherk}, given a $B_{th}$, a lower $\frac{\theta_{k}}{A_{k}}$ (which also indicates a smaller $\beta_k$ as per \eqref{beta}) implies a higher barrier $\Phi_k$ for a participant to benefit. In this case, only if others contribute sufficient data, participants with small $\beta_{k}$ would participate in FL.
Based on the above discussion, we formally derive participant's best response strategy as follows.

\begin{eqnarray}\label{Cbre}
B_{k}^{\ast}=\left\{
\begin{aligned}
&B_{k}^{max}, \qquad \mbox{if} \ B_{k}^{max}<\tilde{B_{k}}.\\
&\tilde{B_{k}},  \qquad \quad \, \mbox{if} \  B_{th} \leq \tilde{B_{k}}, \leq B_{k}^{max}.\\
&B_{th}, \qquad \ \ \   \mbox{if} \ \tilde{B_{k}}<B_{th} \, \mbox{and} \, B_{-k}\geq \Phi_{k}.\\
&0, \qquad \quad \ \ \   \mbox{if} \ \tilde{B_{k}}<B_{th} \, \mbox{and} \, B_{-k}< \Phi_{k}.
\end{aligned}
\right.
\end{eqnarray}
And, $f_{k}^{\ast}$ and $\tilde{B_{k}}$ are obtained by \eqref {fopt} and \eqref{Bpai} respectively.

Similar to COFL, we can define four types of participants as \textit{type-1} ($P_{1}$), \textit{type-2} ($P_{2}$), \textit{type-3} ($P_{3}$) and \textit{type-4} ($P_{4}$) participants in Nash equilibrium, which correspond to the first, second, third and fourth conditions in \eqref{Cbre}.

\subsection{Equivalent form of Nash Equilibrium}
In this subsection, we focus on deriving the characteristics of Nash equilibrium (Theorem \ref{thm4}), based on which we give the equivalent form of Nash equilibrium (Theorem \ref{thm5}).

Similar to COFL, all participants have been sorted in descending order by $\beta_{k}$.
Theorem \ref{thm4} describes the characteristic of Nash equilibrium in CAFL.
\begin{thm}\label{thm4}
	\textit{If CAFL reaches a Nash equilibrium, the participants can be divided into type-1, type-2, type-3 and type-4. For any $i\in P_{1}, c\in P_{2}, j\in P_{3}, k \in P_{4}$, we have $\beta_{k}<\beta_{j}<\beta_{c}<\beta_{i}$. Moreover, at most one participant belongs to type-2 participant.}
\end{thm}

\begin{figure}[t]
	\setlength{\abovecaptionskip}{-0.05cm}
	\centering
	\includegraphics[height=0.25\linewidth]{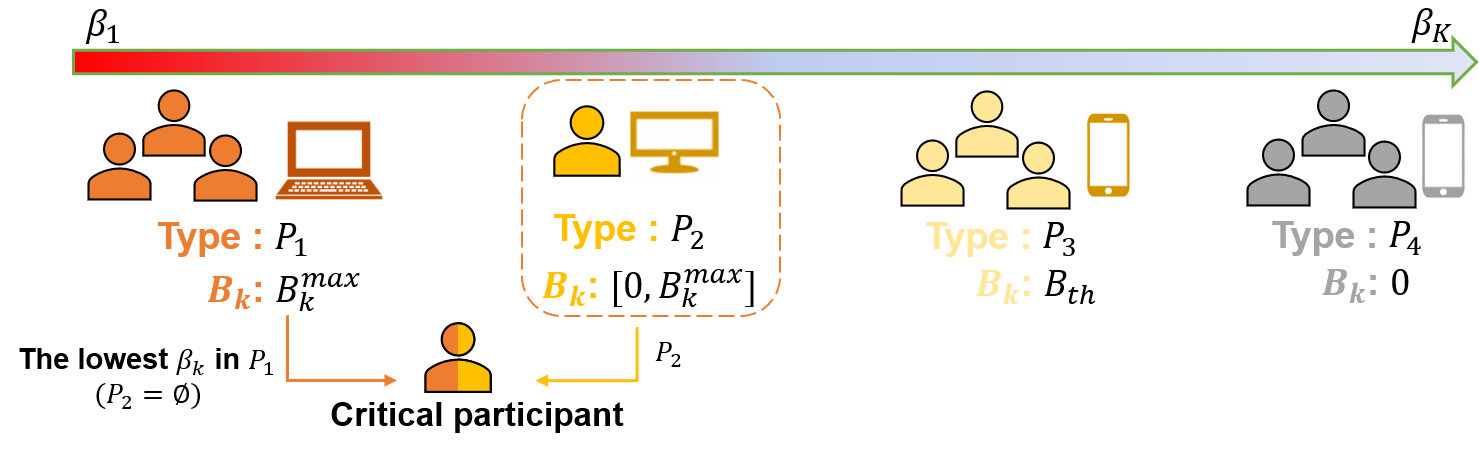}
	\caption{The structure properties of Nash equilibrium in the CAFL game.}
	\label{CAFLNE}
	\vspace{-0.3cm}
\end{figure}

The proof is given in Subsection B in Appendix B in the separate supplementary file. As illustrated in Fig. \ref{CAFLNE}, Theorem \ref{thm4} shows that when the CAFL game reaches the Nash equilibrium, $P_{1}$, $P_{2}$, $P_{3}$ and $P_{4}$ are arranged in descending order of $\beta_{k}$. Similarly, we describe the boundary between $P_{1}$ and $P_{3}$ and the critical participant and equilibrium structure are redefined as:

\noindent\textbf{Definition 3}: Participant $c$ is a critical participant in CAFL game if and only if $\forall i\in \mathcal{K}, \beta_{i}<\beta_{c}$, $\tilde{B_{i}}=\beta_{i}-B_{-i}< B_{th}$ and  $\tilde{B_{c}}=\beta_{c}-B_{-c}\geq B_{th}$ when the game reaches a Nash equilibrium (if it exists).

\noindent\textbf{Definition 4 (participant's equilibrium structure)}: Given a participant $c \in \mathcal{K}$, \textit{\textbf{(i)}} $\forall i \in \mathcal{K},\beta_{i}>\beta_{c}$, $B_{i}=B_{i}^{max}$;
\textit{\textbf{(ii)}} $\forall j \in \mathcal{K}$, $\beta_{j}<\beta_{c}$, $B_{-j} \geq \Phi_{j}$, we set $B_{j}=B_{th}$;
\textit{\textbf{(iii)}} $\forall l \in \mathcal{K}$, $\beta_{l}<\beta_{c}$, $B_{-l} < \Phi_{l}$, we set $B_{l}=0$. We call the strategy profile ($B_{1},...,B_{c-1},B_{c+1},...,B_{K}$) as the participants' equilibrium structure in the CAFL game, where $B_{i} (i\neq c)$ satisfy \textit{\textbf{(i)}}, \textit{\textbf{(ii)}} and \textit{\textbf{(iii)}}.

Based on Theorem 4, we summarize the conditions of one participant becoming the critical participant in CAFL and derive equivalent form of Nash equilibrium in Theorem \ref{thm5}:
\begin{thm}\label{thm5}
	\textit{If there exists a participant $c$ with equilibrium structure, then the CAFL game satisfies one of three properties:}
	\begin{itemize}
		\item {\verb|Property 3|}: $B_{c}^{\ast}=B_{c}^{max}$, $\beta_{c+1}-B_{th} <B_{nash} <\beta_{c}$ and $B_{c+1}=0$.
		\item {\verb|Property 4|}: $B_{c}^{\ast}=B_{c}^{max}$,  $\beta_{c+1}<B_{nash} <\beta_{c}$ and $B_{c+1}=B_{th}$.
		\item {\verb|Property 5|}: $B_{c}^{*} \neq 0$ and $B_{nash}=\beta_{c}$.
	\end{itemize}
	\textit{Where $B_{nash}$ is the global batchsize (i.e., total batchsize contributed by the participants)  at the equilibrium. In this case, the Nash equilibrium exists in CAFL game, and the participant $c$ is a critical participant. Also, if the CAFL game reaches a Nash equilibrium (if it exists), one of the three properties above is satisfied.}
\end{thm}

The proof is given in Subsection C in Appendix B in the separate supplementary file.
The equivalent form of Nash equilibrium motivates us to compute Nash equilibrium using the rules of \textit{\textbf{(i)}}, \textit{\textbf{(ii)}} and \textit{\textbf{(iii)}} in Definition 4 until the one of three properties holds. Theorem \ref{thm5} shows the relation between global batchsize and participants' $\beta_{k}$ at the equilibrium when $c$ is critical participant. Accordingly, we define the global batchsize function at the equilibrium as:
\begin{align}
F_c(c,&B_{c}^{con})=\sum_{k=1}^{c-1}B_{k}^{max}+B_{c}^{con}+\max\limits_{\mathbb{B}}(\sum_{j=c+1}^{K}B_{j}), \label{Fclose}
\\ &\mbox{s.t.}\quad c \in \mathcal{K},  \tag{\ref{Fclose}{a}} \label{Fclosea}
\\ &\qquad B_{th}\leq B_{c}^{con}\leq B_{c}^{max}, \tag{\ref{Fclose}{b}} \label{Fcloseb}
\\ & \qquad B_{j}=B_{th}, \ \mbox{if}\  B_{-j}\geq \Phi_{j} \tag{\ref{Fclose}{c}}, \label{Fclosec}
\\ & \qquad B_{j}=0, \ \mbox{if}\  B_{-j}<\Phi_{j} \tag{\ref{Fclose}{d}} \label{Fclosed}.
\end{align}
Here, $\mathbb{B}=(B_{c+1},...,B_{K})$. The first term of \eqref{Fclose} corresponds to rule \textit{\textbf{(i)}}. The second term is the critical participant $c$'s batchsize constrained by \eqref{Fcloseb}. The third term is constrained by \eqref{Fclosec} and \eqref{Fclosed} originating from rules \textit{\textbf{(ii)}} and \textit{\textbf{(iii)}} respectively.
Given a pair of $c$ and $B_{c}^{con}$, the summation of third term of \eqref{Fclose} may correspond to multiple values. To obtain the equilibrium with the largest batchsize, we choose the maximum global batchsize at the equilibirum as our solution. Since we need to consider the relation between global batchsize and $\beta_{k}$, for convenience, we express three properties in Theorem \ref{thm5} in the equivalent forms using the global batchsize function $F_{c}$ as follows:
\begin{itemize}
	\item{\verb|Property 3|}: $\beta_{c+1}-B_{th}<F_{c}(c, B_{c}^{max})<\beta_{c}$ and $B_{c+1}=0$.
	\item{\verb|Property 4|}: $\beta_{c+1}<F_{c}(c, B_{c}^{max})<\beta_{c}$ and $B_{c+1}=B_{th}$
	\item{\verb|Property 5|}: There exists a $B_{c}^{con}\in [B_{th}, B_{c}^{max}]$ such that $F_{c}(c, B_{c}^{con})=\beta_{c}$.
\end{itemize}
Thus, we can search the equilibrium by checking above properties.

However, Nash equilibrium does not necessarily exist in CAFL game. Table \ref{table_example} shows an example without Nash equilibrium. We obtain $F_c(1,B_{th})=20<\beta_{1}<F_c(1,B_{1}^{max})=100+20$ \footnote{Since $B_{1}=B_{1}^{max}=100$, the global batchsize satisfies $B>\beta_{1}$, which means Property 3 or Property 4 does not hold.}.  We calculate $\Phi(B_{th}, \dfrac{\theta_{2}}{A_{2}})=34.97$. That is, the participant $2$ participates with $B_{2}=20$ in FL if and only if participant $1$'s batchsize is greater than $34.97$. Obviously, we fail to find a $B_{1}$ satisfying $F_c(1,B_{1})=\beta_{1}$, i.e., the Property 5 does not hold. Absence of Nash equilibrium indicates that we can not compute Nash equilibrium as the same as that in the COFL game. In what follows, we will first introduce a partial form of CAFL game, and based on which, we compute Nash equilibrium and address the CAFL game without Nash equilibrium.

\begin{table}[t]
	\setlength{\abovecaptionskip}{-0.02cm}
	\renewcommand{\arraystretch}{1.3}
	\caption{Example without Nash equilibrium}
	\label{table_example}
	\centering
	\scriptsize
	\begin{tabular}{c|c|c|c|c|c}
		\hline
		& $\theta_{k}$ & $A_{k}$& $\beta_{k}$ &$B_{k}^{max}$ &$B_{th}$ \\
		\hline
		\bfseries Participant 1 & 103.41 &1&45.00 & 100& 20\\
		\hline
		\bfseries Participant 2 &9.39 & 1 &2.97 & 100& 20\\
		\hline
	\end{tabular}
	\vspace{-0.3cm}
\end{table}

\subsection{Nash Equilibrium for Partial Form of CAFL Game}
In this subsection, we first consider the case that the pair of $c$ and $B_{c}^{con}$ is given, and accordingly define a partial form of the CAFL game consisting of the participants in set of $\left\lbrace c+1,...,K \right\rbrace$. We prove that such a partial game must possess Nash equilibria.

Specifically, when given a $c$ and $B_{c}^{con}$, we focus on the participants with their $\beta_{k}$ lower than $\beta_{c}$ and consider a new partial game among them where participants only have two strategies, i.e., $B_{th}$ and $0$, which is due to the constraints \eqref{Fclosec} \eqref{Fclosed}. To sum up, the partial form of CAFL game is defined as:
\begin{eqnarray}
G(H,\mathcal{J},\mathcal{S}, U_{j}),
\end{eqnarray}
where
\begin{itemize}
	\item {\it{External parameter}}: $H=\sum_{k=1}^{c-1}B_{k}^{max}+B_{c}^{con}$.
	\item {\it{Participant set}}: $\mathcal{J}=\left\lbrace c+1,...,K\right\rbrace$.
	\item {\it{Participant strategy}}: $\mathcal{S}=\left\lbrace 0,B_{th} \right\rbrace$.
	\item {\it{Utility function}}: $U_{k}$ is defined as:
	\begin{eqnarray}
	U_{j} ( B_{j} , f_{j}^{\ast} )=\theta_{j}\ln(1+\sqrt{H+B_{-j}+B_{j}})- A_{j}B_{j}.
	\end{eqnarray}
\end{itemize}
Here, $B_{-j}=\sum_{i\in\mathcal{J} \setminus\left\lbrace j\right\rbrace }B_{i}$ and $U_j(0,f_{j}^{\ast})=0$. Note that in the partial game $G$, we introduce the external parameter $H$ indicating that participants in $\left\lbrace 1,...,c \right\rbrace$ outsize the partial game $G$ that adopt the given equilibrium strategies of $B^{max}_k$ and $B^{con}_c$. Similar to Proposition 2, the participants' optimal strategies in game $G$ can be determined by the following condition:
\begin{eqnarray}\label{Bgother}
B_{-j} \geq \ \Phi ( B_{th},  \frac{\theta_{j}}{A_{j}} )-H.
\end{eqnarray}
According to the third term of the global batchsize  function $F_c(c,B_{c}^{con})$ in \eqref{Fclose}, we would like to choose the equilibrium with the largest global batchsize $B_{G}=\sum_{i\in\mathcal{J}}B_{i}$ of the partial form game $G$.
Our method to achieve Nash equilibrium with the largest global batchsize in $G$ is given in Algorithm \ref{method1}. The key idea of Algorithm \ref{method1} is to set all participants' strategies to $B_{th}$ initially. And then, it traverses each participant in reverse order to find the first participant who satisfies the condition in \eqref{Bgother}.

\begin{algorithm}[t]
	\caption{Partial Game Nash Equilibrium Computing Algorithm}\label{method1}
	\begin{algorithmic}[1]
		\Require $G(H,\mathcal{J},\mathcal{S}, U_{j})$
		\Ensure Strategy profile $(B_{c+1},B_{c+2},...,B_{K})$
		\State Sort participants in ascending order by $\Phi_{c+1}<...<\Phi_{K}$.
		\State Initially set $\forall j \in \mathcal{J} $, $B_{j}=B_{th}$.
		
		\For {$j=K$ to $c+1$}
		\If {$B_{-j}< \Phi ( B_{th},  \dfrac{\theta_{j}}{A_{j}} )-H$}
		\State $B_{j}=0$.
		\Else
		\State break.
		\EndIf
		\EndFor
		\State \Return $(B_{c+1},B_{c+2},...,B_{K})$
	\end{algorithmic}
\end{algorithm}

\begin{thm}\label{thm8}
	\textit{Partial game $G$ must admit Nash equilibria. Algorithm \ref{method1} can achieve a Nash equilibrium with largest global batchsize with complexity of $O(|\mathcal{J}|)$}.
\end{thm}

The proof is given in Subsection D in Appendix B in the separate supplementary file.
Hence, in the rest of paper, we use the Algorithm \ref{method1} to determine the equilibrium solution for the game $G$. Besides, given $c$ and $B_{c}^{con}$, we can calculate the value of $F_{c}(c,B_{c}^{con})$ through Algorithm \ref{method1}.

\subsection{Nash Equilibrium for Complete Form of CAFL Game}
We next consider the equilibrium solutions of the complete form of CAFL game by all the participants.
According to Theorem \ref{thm5}, if we can search a critical participant $c$ such that one of the properties in Theorem \ref{thm5} is satisfied, then CAFL must have a Nash equilibrium. Otherwise, there would exist some \textit{special participants}, who are not \textit{type-3} and \textit{type-4} but in between causing instable behaviors in the CAFL game, which would impede forming stable collaboration in participant-centric FL. We hence remove these special participants one by one until we can compute a Nash equilibrium.

\begin{figure}[t]
	\setlength{\abovecaptionskip}{-0.05cm}
	\centering
	\includegraphics[width=\linewidth]{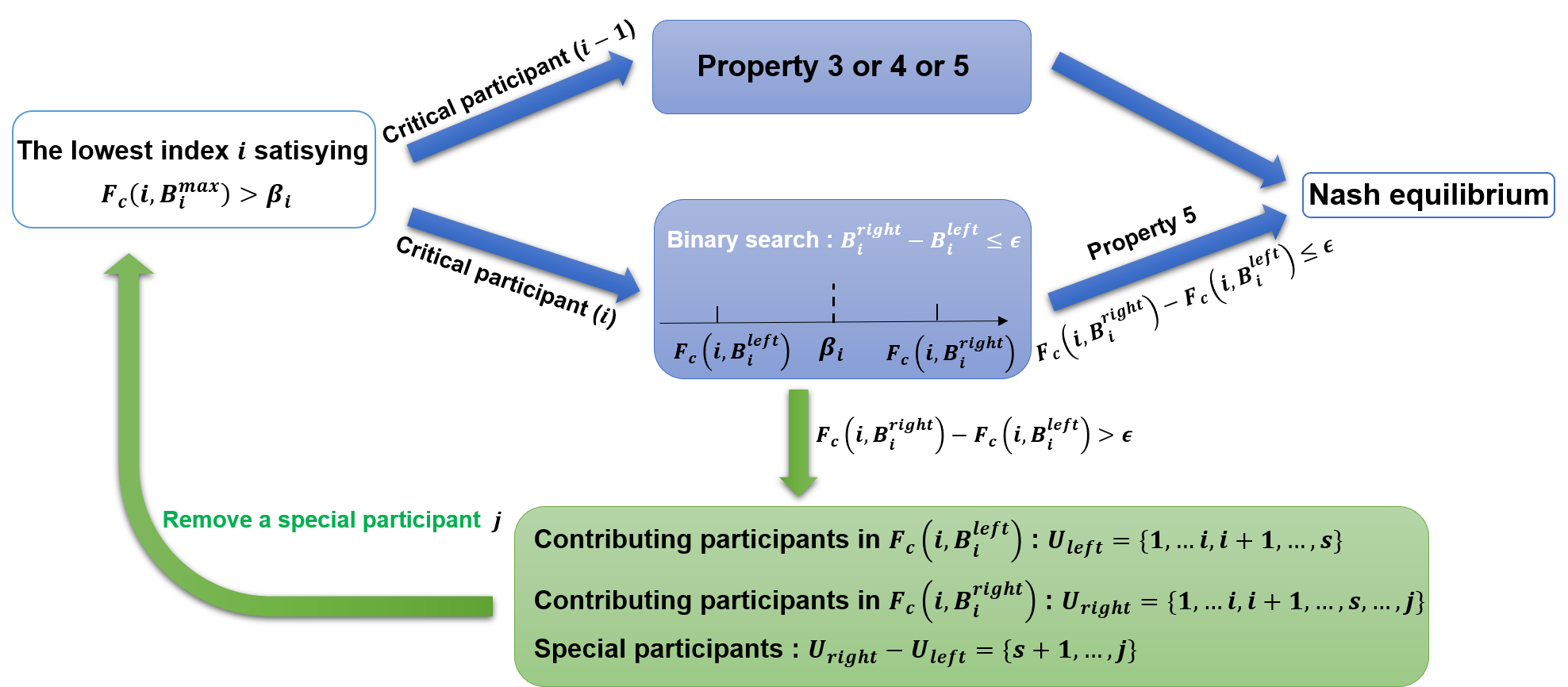}
	\caption{The procedure of computing Nash equilibrium of CAFL game.}
	\label{algorithm3}
	\vspace{-0.2cm}
\end{figure}

The procedure of computing Nash equilibrium is illustrated in Fig. \ref{algorithm3}.
Specifically, based on definition of the global batchsize function $F_c$ in \eqref{Fclose}, $F_c(c,B_{c}^{max})$ is increasing with respect to $c$. We hence first search the lowest participant index $i$ satisfying $F_c(i,B_{i}^{max})>\beta_{i}$, which indicates $F_c(i-1,B_{i-1}^{max})\leq \beta_{i-1}$. Then, \textbf{the possible critical participant must be $i-1$ or $i$}.

\begin{itemize}
\item {\it{$i-1$ as the critical participant}}:
we then first determine whether $i-1$ is a critical participant by checking the relationship among $F_c(i-1,B_{i-1}^{max})$, $\beta_{i-1}$ and $\beta_{i}$ to verify whether Property 3 or 4 or 5 holds.

\item {\it{$i$ as the critical participant}}:
Otherwise, when $i$ is a critical participant, Property 3 or 4 does not hold, since $F_c(i,B_{i}^{max})>\beta_{i}$.
We then attempt to search a $B_{i}^{con}\in [B_{th},B_{i}^{max}]$ satisfying $F_c(i,B_{i}^{con})=\beta_{i}$ to verify whether Property 5 holds with $i$ as the critical participant.
Note that, $F_{c}(i,B_{i}^{con})$ is also increasing with respect to $B_{i}^{con}$, since larger first and second terms of $F_c$ (larger $H$ in game $G$) means larger third term.
Thus, Property 5 ($i$ as critical participant) indicates that $F_c(i,B_{th})\leq\beta_{i}< F_c(i,B_{i}^{max})$.
This motivates us to process \textit{binary search} to finish searching. We predefine a small threshold as search accuracy $\epsilon$.
When the Property 5 holds ($i$ as critical participant), there exists a $B_{i}^{con}\in [B_{th},B_{i}^{max}]$ satisfying $F_c(i,B_{i}^{con})=\beta_{i}$. We will obtain the result:
$B_{right}-B_{left}\leq \epsilon$ and $F_c(i,B_{right})-F_c(i,B_{left})\leq \epsilon$. We can choose one of $B_{left}$ and $B_{right}$ as $i$'s strategy. Here, $B_{left}$ and $B_{right}$ are boundaries of the interval of binary search.

\item {\it{Nash equilibrium dose not exist}}:
when the Property 5 does not hold, we will fail in searching a $B_{i}^{con}\in[B_{th},B_{i}^{max}]$ satisfying $F_{c}(i,B_{i}^{con}) = \beta_{i}$. In this case, the CAFL game does not have a Nash equilibrium. As a result, we will try to remove some special participants in the game such that the refined CAFL game has a Nash equilibrium.
Intuitively, these special participants can be plausible contributing participant candidates but cannot satisfy Property 5 exactly. In this case, they are very sensitive to minor changes of other participants' decisions, causing the instable equilibrium behavior of switching between contributing ($B_i= B_{th}$) and quiting ($B_i=0$) after the best response adjustments by others.
Formally, the above special participants' characteristic will cause the \textit{binary search} stopping with following results:
$B_{right}-B_{left}\leq \epsilon$ and $F_c(i,B_{right})-F_c(i,B_{left})>\epsilon$, with $F_c(i,B_{left})<\beta_{i}<F_c(i,B_{right})$. This indicates that $F_c(i,B_{i}^{con})$ is not a continuous function respect to $B_{i}^{con}$ in the small interval $B_{i}^{con}\in [B_{left},B_{right}]$.
According to the global batchsize function $F_c(i, B_{left})$ in \eqref{Fclose}, we denote $U_{left}$ as the contributing participant set in which all the participants' strategy batchsizes is non-zero. Similarly, the contributing participant set $U_{right}$ corresponds to $F_c(i,B_{right})$.
The set of special participants can be obtained by $U_{right}-U_{left}$. We can then remove a special participant with the lowest $\beta_{k}$ and repeat searching the critical participant until we succeed in achieving Nash equilibrium.
\end{itemize}

Note that when (refined) CAFL game has a Nash equilibrium with the identified critical participant $c$, we can compute the equilibrium strategies as follows: 1) we set $B_k=B^{max}_{k}, \forall k\in\{1,...,c-1\};$ 2) $ B_c=B^{con}_c$; 3) obtain the strategies for the participants in the set $\{c+1,...,K^{new}\}$ using Algorithm 2 based on the partial game $G$. Here, $K^{new}$ denotes the total number of participants in the refined CAFL game after removing the special participants if the original CAFL game does not have a Nash equilibrium. For the removed participants, we have $B_k=0$ and they are excluded from obtaining the trained FL model.

Due to space limit, more analysis on the above equilibrium construction procedure is detailed in Subsection E in Appendix B in the separate supplementary file.

\begin{algorithm}[t]
	\caption{CAFL Nash Equilibrium Computing Algorithm}
	\label{close}
	\begin{algorithmic}[1]
		\Require $K$ participants, batchsize threshold $B_{th}$
		\Ensure $(B_{1},B_{2},...,B_{K})$
		\State Search the lowest participant's index $i$ satisfying $F_c(i,B_{i}^{max})> \beta_{i}$.
		
		\State  Check Property 3 or 4 or 5 ($i-1$ as the critical participant) through the relationship among $F_c(i-1,B_{i-1}^{max})$, $\beta_{i-1}$ and $\beta_{i}$.
		\If {$i-1$ is the critical participant}
		\State Construct equilibrium strategies by setting $c=i-1$.
		\State Goto Line 16.
		\Else
		\State Binary search a $B_{i}^{con}\in [B_{th},B_{i}^{max}]$ satisfying $F_c(i,B_{i}^{con})=\beta_{i}$.
		\If {succeed in searching $B_{i}^{con}$}
		\State Construct equilibrium strategies by setting $c=i$.
		\State Goto Line 16.
		\Else
		\State Remove the special participant with highest index $j$.
		\State Goto Line 1.
		\EndIf
		\EndIf
		\State \Return equilibrium strategies $(B_{1},B_{2},...,B_{K})$
	\end{algorithmic}
	
\end{algorithm}

\section{CAFL Game based Algorithm Design and implementation}
In this section, we summarize the results above to form a CAFL game based collaboration strategy for participant-centric FL. We further discuss the implementation and potential problems for the proposed mechanism in realistic deployment.

\subsection{Algorithm Design}
We first propose our approach to compute a Nash equilibrium for the CAFL game in Algorithm \ref{close}, with the key idea of searching the correct critical participant. In Line 1, we first search the critical participant ($i-1$ or $i$) using the binary search. In Lines 2 and 3, if $i-1$ is the critical participant at equilibrium, his strategy must be $B_{i-1}^{max}$. In Lines 6 and 7, we consider the case that $i$ is the critical participant. Line 8 corresponds to Property 5 ($i$ as critical participant). In Line 12, when Property 5 does not hold, we fail in achieving Nash equilibrium. Only one participant with the lowest $\beta_{k}$ in the special participants (obtained in binary search) is removed. In Line 13, we repeat all steps and go to Line 1.
We can characterize the convergence of Algorithm \ref{close} in Theorem \ref{thm7}.

\begin{thm}\label{thm7}
	\textit{In Algorithm \ref{close}, the maximum number of iterations (i.e., executions of line 13) is $K-i^{\ast}$, where $i^{\ast}$ is the lowest index $i$ satisfying $F_c(i,B_{i}^{max})> \beta_{i}$ in the original CAFL game.
		The solution searched by Algorithm \ref{close} is a Nash equilibrium for the refined CAFL game of the set of the remaining participants.}
\end{thm}

The proof is given in Subsection F in Appendix B in the separate supplementary file. We further analyze the complexity of Algorithm \ref{close}. We first obtain the complexity for one iteration (Line 1-12). The complexity of calculating $F_c(i,B_{i}^{max})$ is $O(K)$. Hence, the complexity of Line 1 is $O(K\log(K))$. In Line 4, constructing equilibrium strategy for each participant, the complexity is $O(K)$. In Line 7, the complexity of binary search is $O(\log(\frac{B^{max}-B_{th}}{\epsilon}))$.
Here, $B^{max}=\underset{\forall k \in \mathcal{K}}{\max}(B_{k}^{max})$.
Thus, the complexity of one iteration is $O(K\log(\frac{B^{max}-B_{th}}{\epsilon})+K\log(K))$.
Based on Theorem 7 (the maximum number of iteration is bounded by $K$), the complexity of Algorithm \ref{close} is  $O(K^2\log(\frac{B^{max}-B_{th}}{\epsilon})+K^2\log(K))$. Note that, a lower $\epsilon$ leads to a higher complexity.

We next consider boosting the global efficiency of the CAFL game by searching  the optimal threshold batchsize $B_{th}\in [1,B_{th}^{max}]$ to maximize the total utility. Specifically, we define the total utility of all in equilibrium defined as:

\begin{eqnarray}\label{Tporige}
TU= \sum_{k\in\mathcal{U} }\left( \theta_{k} \ln(1+\sqrt{B_{-k}+B_{k}})- A_{k}B_{k}\right).
\end{eqnarray}
Here, $\mathcal{U}$ represents the contributing participant set in which all the participants' strategy batchsizes is non-zero in Nash equilibrium. For practical implementation, we can only consider the choices of discrete minimum threshold batchsize, i.e., $B_{th}=1,2,...,B_{th}^{max}$.
Given a $B_{th}$, the total utility can be calculated by performing Algorithm \ref{close} to obtain the equilibrium solution. Thus, we can obtain the optimal threshold batchsize through exhaustive search.
The running time for searching the optimal threshold batchsize is illustrated in Table \ref{table_linear}. $B_{th}^{max}$ indicates the number of execution of Algorithm \ref{close}.
As for average execution time of Algorithm \ref{close},
for example, given a fixed $B_{th}^{max}=100$ and $K=100$, the average execution time of Algorithm \ref{close} (determine a Nash equilibrium points) is $\frac{0.5193}{B_{th}^{max}}=0.05193s$, which indicates the high computational efficiency of Algorithm \ref{close}.
Note that running time for searching the optimal threshold batchsize is short in general, the average running time (repeated over 20 times) is less than 56 seconds when the number of participants $K=1000$ and the maximum batchsize $B_{th}^{max}=1000$, which is negligible compared with the time consuming FL model training process.

\begin{table}[t]
	\setlength{\abovecaptionskip}{-0cm}
	\renewcommand{\arraystretch}{1.3}
	\caption{The average running time(s) of optimal threshold searching}
	\label{table_linear}
	\scriptsize
	\centering
	\begin{tabular}{c|c|c|c}
		\hline
		\diagbox [width=8em,trim=l] {$K$}{$B_{th}^{max}$}&100 & 500& 1000\\
		\hline
		100 & 0.5193& 2.270 & 4.374\\
		500 & 3.185 & 15.22 & 28.52\\
		1000 &7.080 & 33.98 &55.31\\
		\hline
	\end{tabular}
\end{table}

\subsection{Discussion on Implementation and Potential Problems}
In practice, for deployment of the CAFL game based incentive mechanism, each participant can first calculate his unit training cost $A_{k}$ locally based on its own information, and then reports the parameters of unit cost $A_{k}$, maximum data size $B_k^{max}$ and model valuation $\theta_{k}$ to the FL server (which plays a neutral role in participant-centric FL).
Based on the participants' reported information, the server will compute the optimal equilibrium strategies with optimal threshold for the participant-centric FL using the algorithms above, and then announce the strategies to the participants. We should emphasize that each participant has incentive to follow the announced strategy due to the property of Nash equilibrium and the enforcement of contribution threshold mechanism by the FL server.

We have achieved a lightweight meachnism to collaborate participants to finish the FL task. In what follows, we further discuss the potential problems in realistic deployment.

\textbf{Model Security}: Similar to many existing studies on incentive issues \cite{Auction21} \cite{Contract21} \cite{public21}, we assume that all participants are selfish but have no intention of sabotaging the FL model. In terms of potential model security issues, using validation dataset is a lightweight and economic method to check quality of the uploaded model from each participant in order to defend against data poisoning attack or model manipulation (e.g., dirty-label data \cite{Robust2022}).
In response to the backdoor attack \cite{wang2020attack}, we can refer to three state-of-art defense mechanisms to defend it from the perspective of certified defense \cite{xie2021crfl}, validation datasets detection \cite{baffle21} and robust aggregation \cite{wrobust21}, respectively. These aforementioned mechanisms can be easily applied in our participant-centric FL scenario.

\textbf{Non-IID Data}: we assume that all participants have the common interest and would like to obtain the global trained FL model and their possess IID data, while neglecting the difference of data distribution among participants to some extend (non-IID). From economic perspective,
non-iid case will not change the essential characteristic of COFL (free-riding phenomenon). The non-iid issue will be considered in future work. Intuitively, one potential extension is to adjust the participants' overall model valuation parameters, since the overall non-iid level in a large-scale group is relatively static.

\textbf{Communication or computation disruption}: we focus on
the participants' collaboration in FL from theoretical perspective in this paper. In realistic deployment, a participant may experience failures in terms of communication or computation disruptions probabilistically. To address such a risk-aware FL collaboration scenario, one study direction is to integrate the prospect theory \cite{2013prospect} \cite{Risk20} with our game model to analyze risk-aware decision making behaviors when facing with uncertainty.
Intuitively, risk-aware decision makings would make
the participants to be more conservative in collaboration. The rigorous analysis on this based on prospect theory is
mathematically involved and out of scope of this study. We will consider it in a future work.

\textbf{Truthful parameter reporting}: the issue of participant's truthful parameter reporting is not the focus of this work and will be considered in a future work. Intuitively, if a participant exaggerates his model valuation, he would be required to contribute more. While, if he understates his valuation excessively, he would risk at getting removed from FL. In practice, as per the prior statistical distributions of participants' valuations, we can globally finetune the contribution threshold to balance the truthfulness and optimality.

\section{NUMERICAL RESULTS}
In this section, we evaluate CAFL performance on global batchsize and total utility in different structures of groups. Experiment results show that CAFL effectively alleviates the ``free-riding'' phenomenon.
We also study the impact of parameters on participants' behaviors. We find that participant's position in group (i.e., the relative size of $\beta_{k}$) has a great influence on the participant's strategy. We will also evaluate the performance of the FL model achieved by CAFL and COFL in the realistic MNIST, FashionMNIST and CIFAR10 datasets.

\subsection{Contribution-Aware FL Performance}\label{pf}

\textit{Simulation setting}: We first divide each group into two types of participants. (i) \textit{High quality (Hq)}: The parameter  $\theta_{k}$ varies from $50$ to $100$. The $f_{k}^{min}$ and $f_{k}^{max}$ are 0.3 GHz and 1.5 GHz. We set parameters $\varphi_{k} \in [1,10]$ and $\gamma_{k} \in [10,100]$. And, $C_{k}$ is uniformly distributed in $[ 1.22\times 10^{6}, 2.44 \times 10^{6}]$; (ii) \textit{Low quality (Lq)}: The parameter $\theta_{k}$ ranges from 0 to 10. We set parameters $\varphi_{k} \in [1,20]$ and $\gamma_{k} \in [10,200]$. Other parameter settings are the same as that of \textit{High quality}. Parts of the above parameter settings are based on \cite{Fed19}. The device for simulation in this paper is equipped with a 8-core Intel(R) Core(TM) i7-8650U CPU and 1 NVIDIA GeForce GTX 1060 GPU. The device for model training (Subection \ref{tr_sec}) is based on Ubuntu 18.04.05, CUDA v11.6 and Intel(R) Xeon(R) CPU (E5-2678 v3).

\subsubsection{The performance of CAFL vs COFL}
In this subsection, we study the performance of CAFL in different structures of participant groups.

We study the results for different groups with different proportions of \textit{High quality} participant under different numbers of participants. Here, $B_{k}^{max} \in [30, 150]$. In order to reduce the experimental error caused by parameters' randomness, we run 100 times under each type of groups and average the results. Note that, $B_{th}$ in CAFL is the group optimal minimum threshold batchsize.

\begin{table}[t]
	\setlength{\abovecaptionskip}{-0cm}
	\renewcommand{\arraystretch}{1.3}
	\caption{The average number of contributing participants in difference groups}
	\label{table_renshu}
	\scriptsize
	\centering
	\begin{tabular}{c|c|c|c|c|c}
		\hline
		\diagbox [width=8em,trim=l] {$K$}{Hq(\%)}& 0 & 25 & 50 & 75 & 100\\
		\hline
		20 (COFL) & 1.13& 2.66 & 3.41& 3.86& 4.26 \\
		20 (CAFL) & \bfseries 18.84 & \bfseries 15.02&\bfseries 17.02& \bfseries 17.27 &\bfseries 20 \\
		50 (COFL) & 1.38 & 3.71 & 4.52& 4.96&5.27 \\
		50 (CAFL) &\bfseries 45.74 &\bfseries 43.28 &\bfseries 45.22& \bfseries 47.69&\bfseries 50 \\
		
		100 (COFL) & 1.53 & 4.57 & 5.28& 5.89& 6.30 \\
		100 (CAFL) &\bfseries 93.47 &\bfseries 87.49 &\bfseries 91.19& \bfseries 96.05&\bfseries 100 \\
		\hline
	\end{tabular}
\end{table}

\begin{figure}[H]
	\centering
	\includegraphics[height=0.25\textwidth]{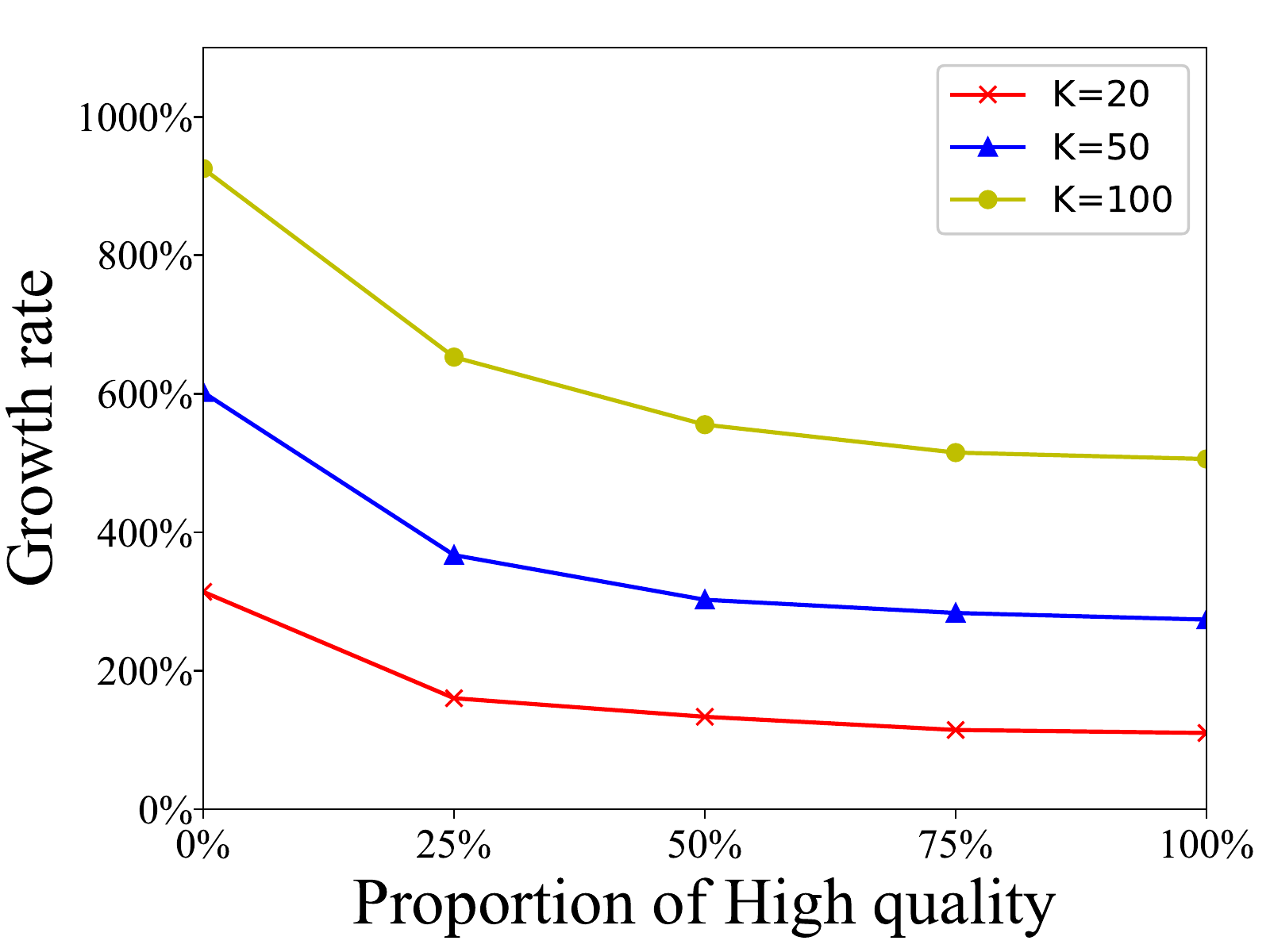}
	\caption{Growth rate of CAFL on the global batchsize.}
	\label{batchup}
\end{figure}
\begin{figure}[H]
	\centering
	\includegraphics[height=0.25\textwidth]{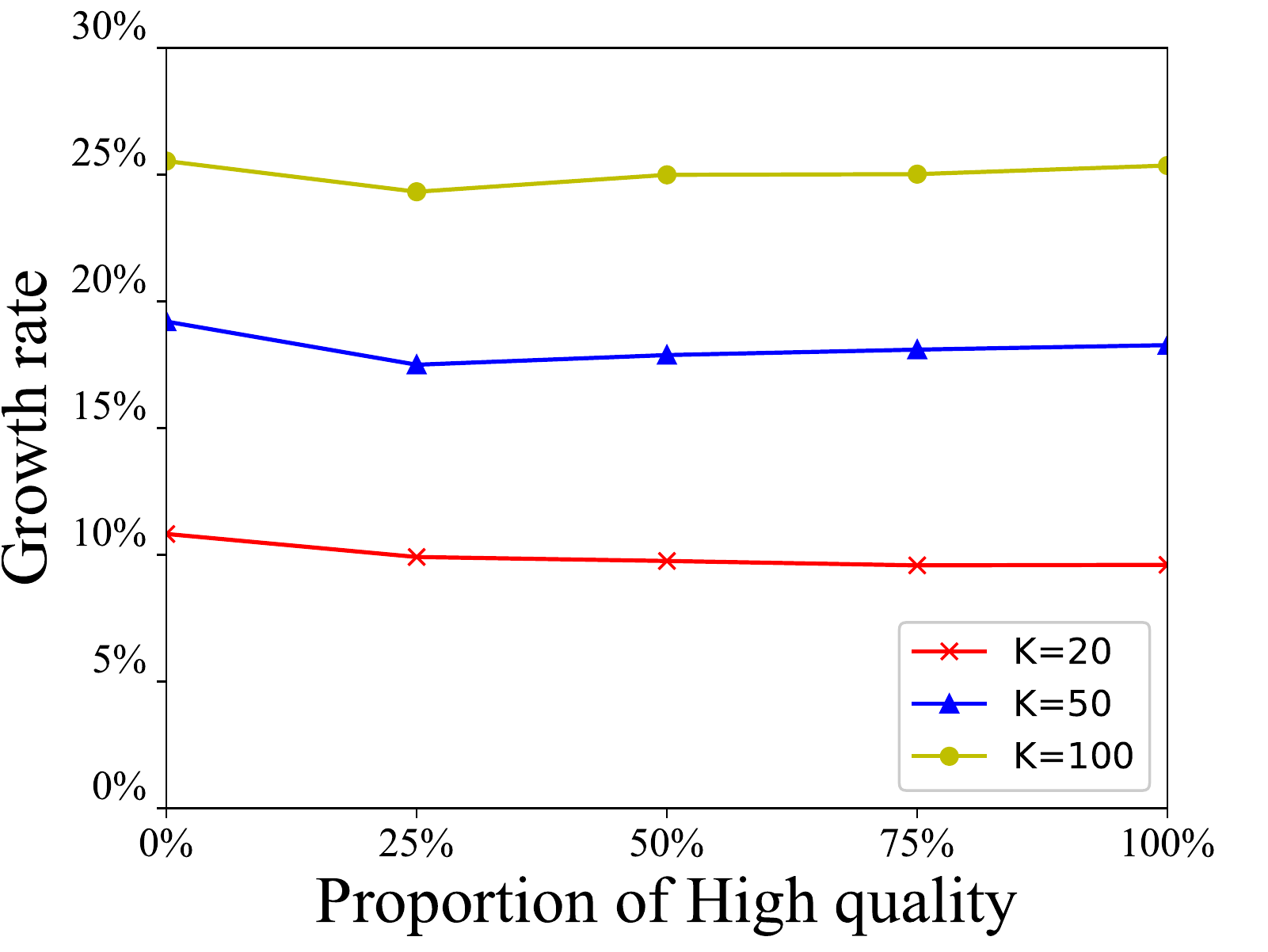}
	\caption{Growth rate of CAFL on the total utility.}
	\label{profitup}
\end{figure}

\begin{figure*}[t]
	\begin{minipage}[t]{0.33\textwidth}
		\setlength{\abovecaptionskip}{-0.05cm}
		\centering
		\includegraphics[width=\linewidth]{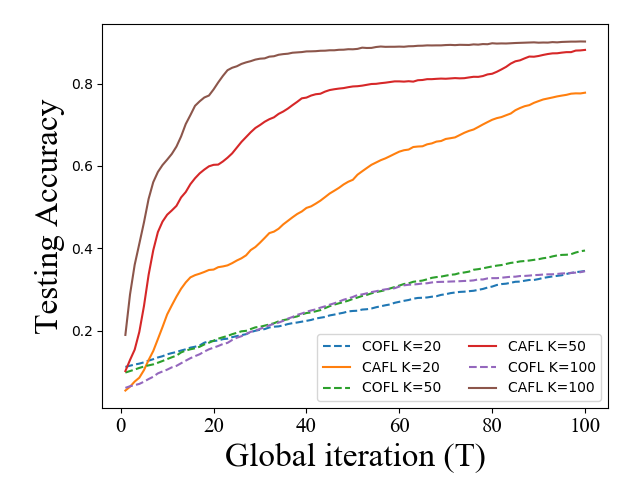}
		\caption{The comparison of accuracy with COFL and CAFL on MNIST dataset}
		\label{ministacc}
	\end{minipage}
	\begin{minipage}[t]{0.33\textwidth}
		\setlength{\abovecaptionskip}{-0.05cm}
		\centering
		\includegraphics[width=\linewidth]{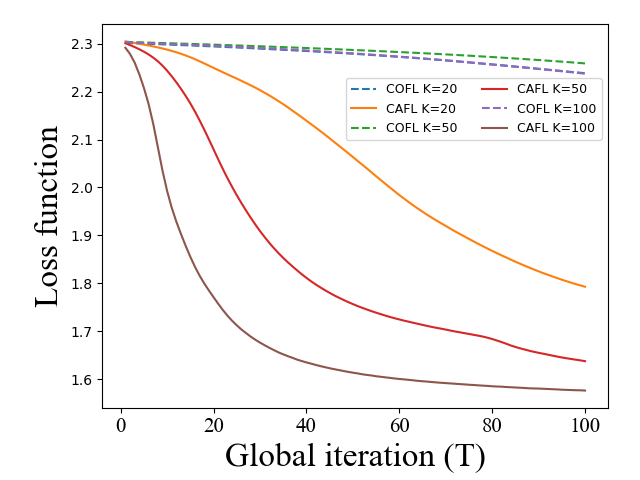}
		\caption{The comparison of loss function with COFL and CAFL on MNIST dataset}
		\label{ministloss}
	\end{minipage}
	\begin{minipage}[t]{0.33\textwidth}
		\setlength{\abovecaptionskip}{-0.05cm}
		\centering
		\includegraphics[width=\linewidth]{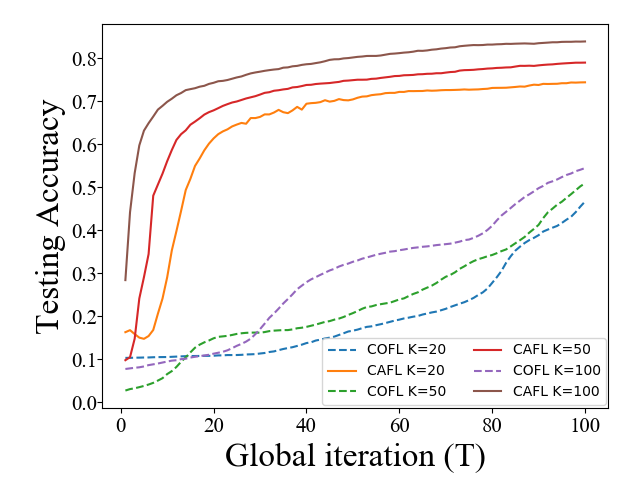}
		\caption{The comparison of accuracy with COFL and CAFL on FMINIST dataset}
		\label{AccFM}
	\end{minipage}
	\vspace{-0.2cm}
\end{figure*}

\begin{figure*}[t]
	\begin{minipage}[t]{0.33\textwidth}
		\setlength{\abovecaptionskip}{-0.05cm}
		\centering
		\includegraphics[width=\linewidth]{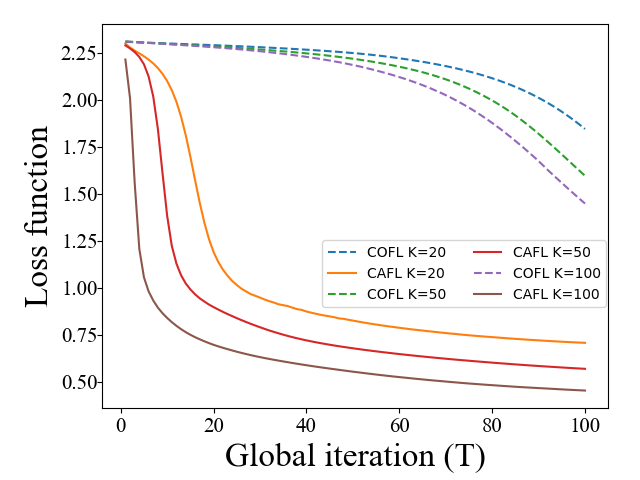}
		\caption{The comparison of loss function with COFL and CAFL on FMINIST dataset}
		\label{LossFminist}
	\end{minipage}
	\begin{minipage}[t]{0.33\textwidth}
		\setlength{\abovecaptionskip}{-0.05cm}
		\centering
		\includegraphics[width=\linewidth]{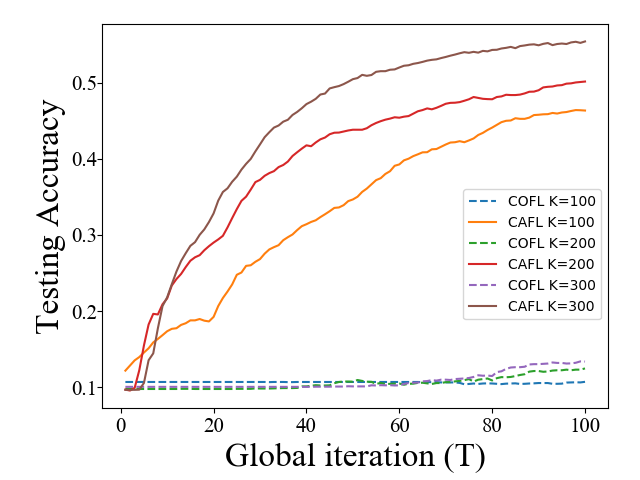}
		\caption{The comparison of accuracy with COFL and CAFL on CIFAR10 dataset}
		\label{cifaracc}
	\end{minipage}
	\begin{minipage}[t]{0.33\textwidth}
		\setlength{\abovecaptionskip}{-0.05cm}
		\centering
		\includegraphics[width=\linewidth]{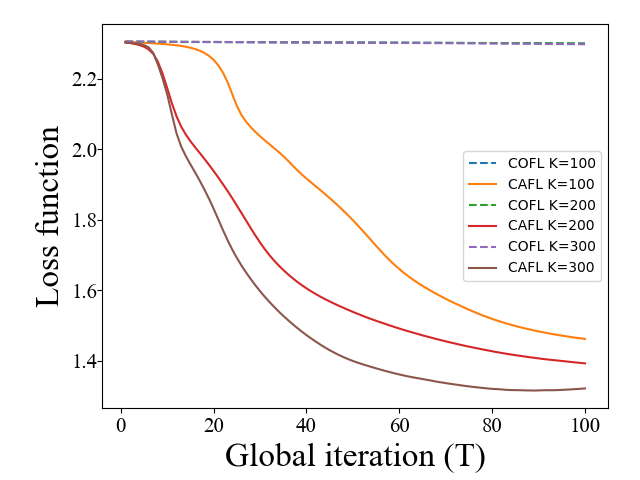}
		\caption{The comparison of loss function with COFL and CAFL on CIFAR10 dataset}
		\label{cifarloss}
	\end{minipage}
	
	\vspace{-0.2cm}
\end{figure*}

As shown in Table \ref{table_renshu}, in the COFL, under the same ratio of \textit{High quality} (\textit{Hq}) participant is unchanged, the proportion of non-contributing participants gradually increases with the number of participants in group. This shows that the larger of participant size in the, the more severe free-riding phenomenon. Besides, there are much more contributing participants in CAFL (more than 90\% participants contributed in most cases) than in COFL, which means CAFL effectively alleviates the free-riding phenomenon.
Furthermore, the number of contributing participants in CAFL with $25\%$ \textit{High quality} is less than that in CAFL with 0\% \textit{High quality}. The reason is as  follows: When the proportion of \textit{High quality} is 0\%, the threshold $B_{th}$ ranges from 9 to 12 when number of participants $K$ equals to 20, 50 and 100, respectively. However, when the proportion of \textit{High quality} is $25\%$, $B_{th}$ ranges from 30 to 35.
The high threshold batchsize $B_{th}$ makes more low-quality participants choose to exit. When the proportion of \textit{High quality} continues to increase from 25\%, the threshold $B_{th}$ does not increase much, leading the increasing number of contributed participants.

Fig. \ref{batchup} and \ref{profitup} show the growth rate of CAFL relative to COFL on the global batchsize and total utility, respectively.
As the number of participants increases, the free-riding phenomenon becomes more and more serious, and hence  performance gain (both on global batchsize and total utility) of CAFL is more superior.
For instance, CAFL can achieve up to 9x growth in terms of the global data batchsize over COFL, which also implies a significant improvement on the FL model accuracy. Moreover, the growth rates of CAFL in both utilities and global batchsize over COFL increase with the number of total participants $K$, which demonstrate that CAFL can be more efficient for large-scale FL applications. Note that for a fixed number of participants $K$, with a larger proportion of \textit{High quality} participants, the impact of free-riding is slightly weaken in COFL, and hence the growth rates of CAFL change smoothly in Fig. \ref{batchup} and \ref{profitup}.

\subsubsection{The training performance of CAFL vs COFL}\label{tr_sec}

We further evaluate the performance of CAFL in MNIST and CIFAR10 dataset, compared with COFL.

\textit{MNIST and FashionMNIST (FMINIST) Setting}: The standard MNIST \cite{MNIST98} and FMNIST \cite{fashion17} consist of 60000 training samples and 10000 test samples. For MNIST setting, we use a multi-layer perception (MLP) network only with one hidden layer (256 hidden unit). For FMNIST setting, we utilize a network with 2 convolutional layers and 1 fully connected layer.
We set learning rate $\eta=0.01$ and local epoch equaling to 1. We conduct the experiment on three groups with different number of participants (50\% \textit{Hq} and 50\% \textit{Lq} participants) in COFL and CAFL respectively. Here, $B_{k}^{max}\in [200,300]$.

\textit{CIFAR10 Setting}: CIFAR10 dataset \cite{cifar10} has 50000 training examples and 10000 test examples. We use LeNet consisting of two sets of convolution and pooling layers, then two fully-connected layers with ReLU activation. The learning rate and local epoch are set to $\eta=0.01$ and $1$ respectively. The participant groups settings are the same as that in \textit{MNIST setting}.

We compare the accuracy and loss function with COFL and CAFL in Fig. \ref{ministacc}, \ref{ministloss}, \ref{AccFM}, \ref{LossFminist}, \ref{cifaracc} and \ref{cifarloss}. In a finite number of global iterations, the  training performance of CAFL is superior of COFL on three datasets, since CAFL can achieves a larger global batchsize FL. In terms of test accuracy ($T=100$), CAFL are 43.26\%, 48.68\% and 55.71\% more than COFL on MNIST dataset in three groups ($K=20$, $K=50$ and $K=100$ respectively). On FMNIST dataset, the accuracy of CALF are 27.75\%, 28.07\%, and 29.47\% more than that of COFL in three groups ($K=20$, $K=50$ and $K=100$ respectively).
Similarly, on CIFAR10 dataset, the accuracy of CAFL ($T=100$) are 35.61\%, 37.65\% and 42.02\% more than that of COFL in three groups ($K=100$, $K=200$ and $K=300$ respectively). This implies that CAFL can greatly improve the training performance with a large number of total participants $K$.

\begin{figure*}[t]
	\vspace{0cm}
	\begin{minipage}[t]{0.33\textwidth}
		\setlength{\abovecaptionskip}{-0.00cm}
		\centering
		\includegraphics[width=\linewidth]{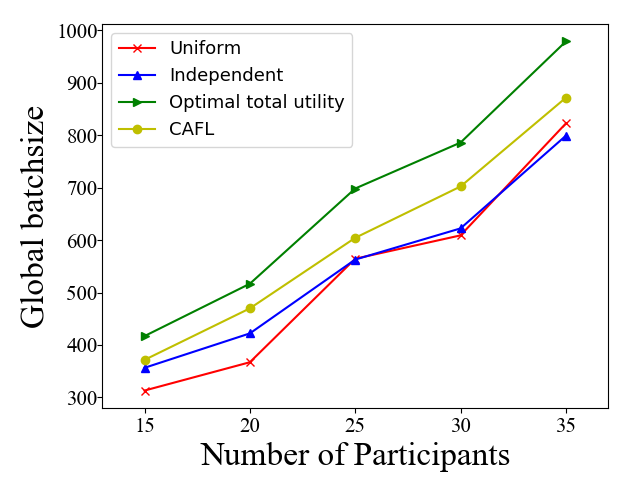}
		\caption{The average global batchsize for each scheme}
		\label{compare_batch}
	\end{minipage}
	\begin{minipage}[t]{0.33\textwidth}
		\setlength{\abovecaptionskip}{-0.05cm}
		\centering
		\includegraphics[width=\linewidth]{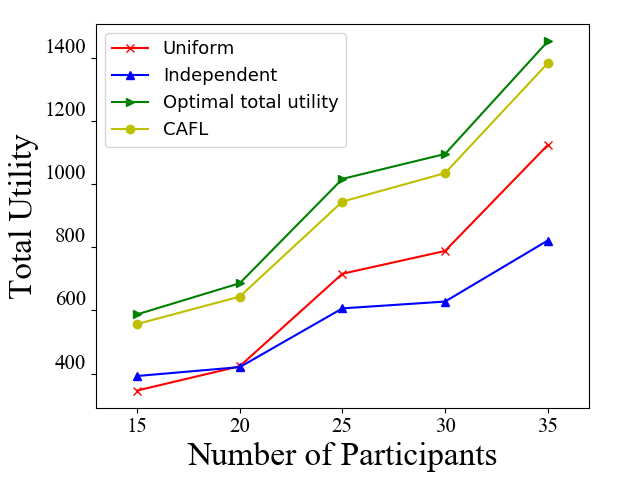}
		\caption{The average total utility for each scheme}
		\label{compare_profit}
	\end{minipage}
	\begin{minipage}[t]{0.33\textwidth}
		\setlength{\abovecaptionskip}{-0.05cm}
		\centering
		\includegraphics[width=\linewidth]{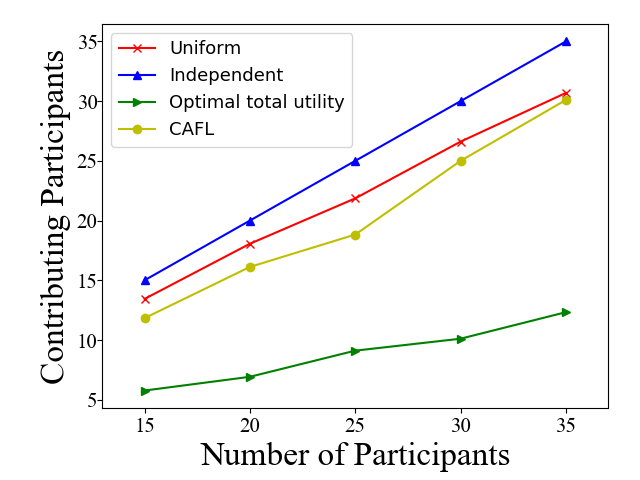}
		\caption{The average contributing participants for each scheme}
		\label{compare_participant}
	\end{minipage}
	\vspace{-0.25cm}
\end{figure*}

\begin{figure*}[t]
	\begin{minipage}[t]{0.24\textwidth}
		\setlength{\abovecaptionskip}{-0.05cm}
		\centering
		\includegraphics[width=\linewidth]{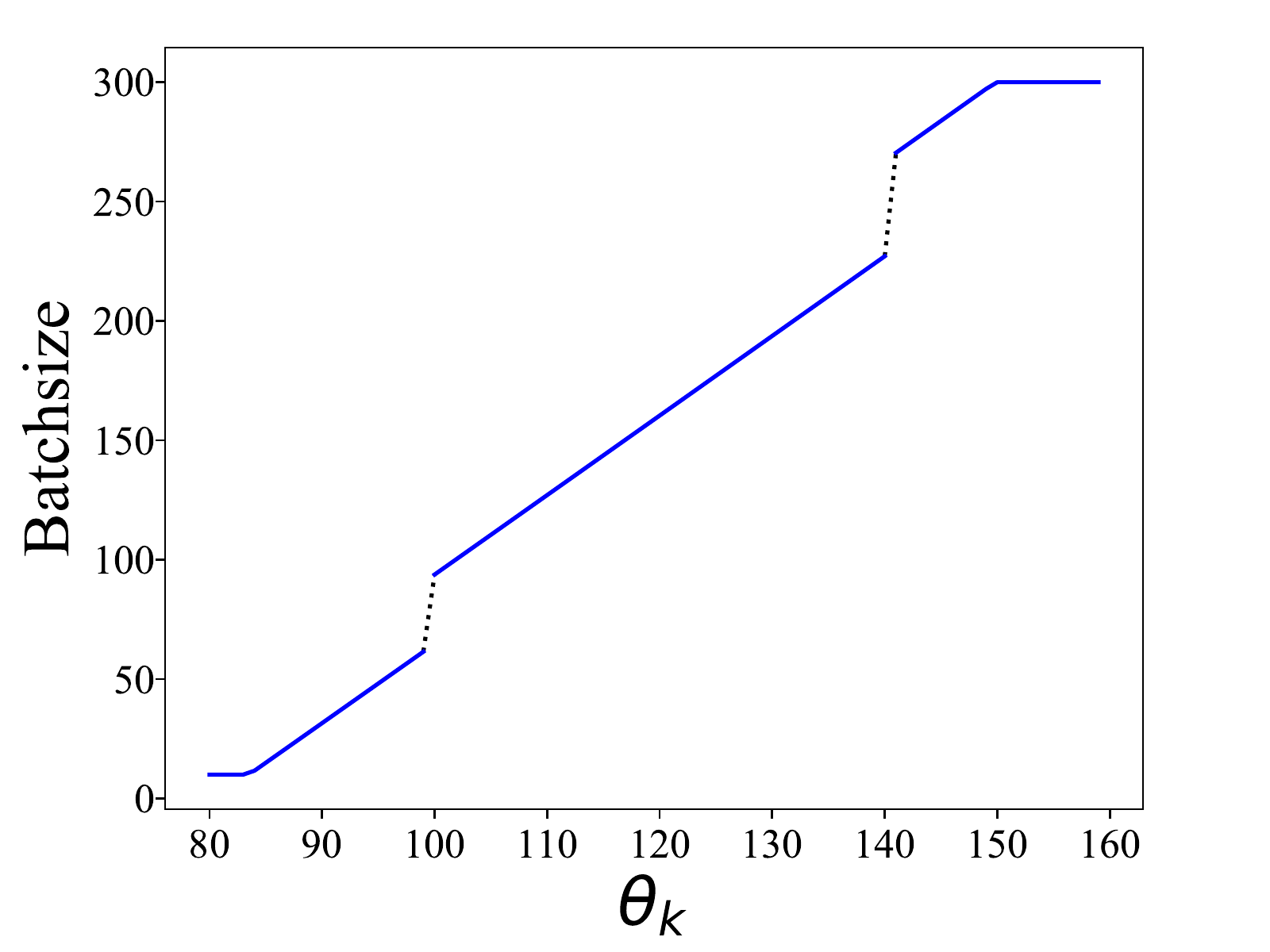}
		\caption{Impact of $\theta_{k}$ on participants' strategies $B_{k}$}
		\label{batch_theta}
	\end{minipage}
	\begin{minipage}[t]{0.24\textwidth}
		\setlength{\abovecaptionskip}{-0.05cm}
		\centering
		\includegraphics[width=\linewidth]{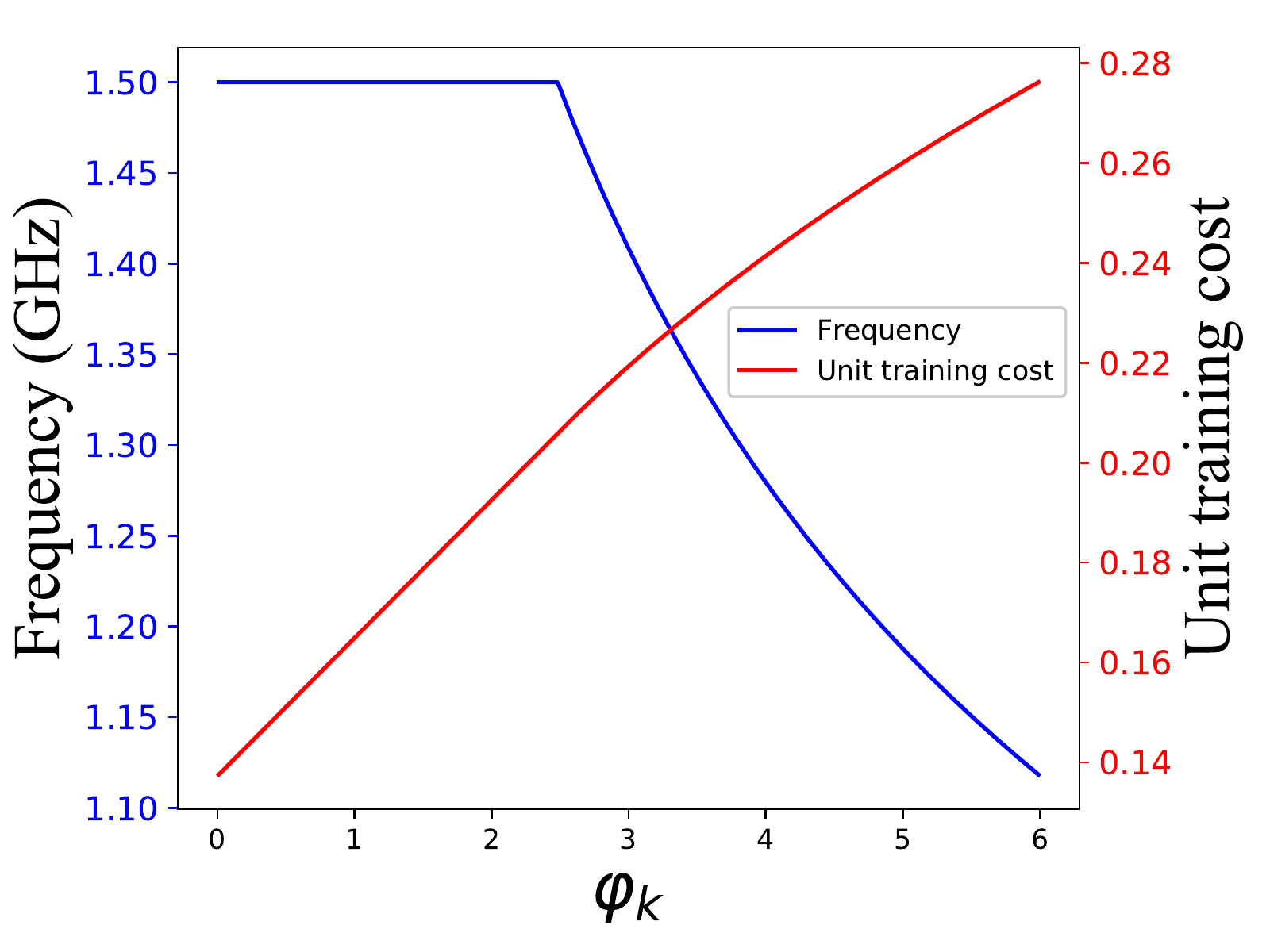}
		\caption{Impact of $\varphi_{k}$ on $f_{k}$}
		\label{batch_faif}
	\end{minipage}
	\begin{minipage}[t]{0.24\textwidth}
		\setlength{\abovecaptionskip}{-0.05cm}
		\centering
		\includegraphics[width=\linewidth]{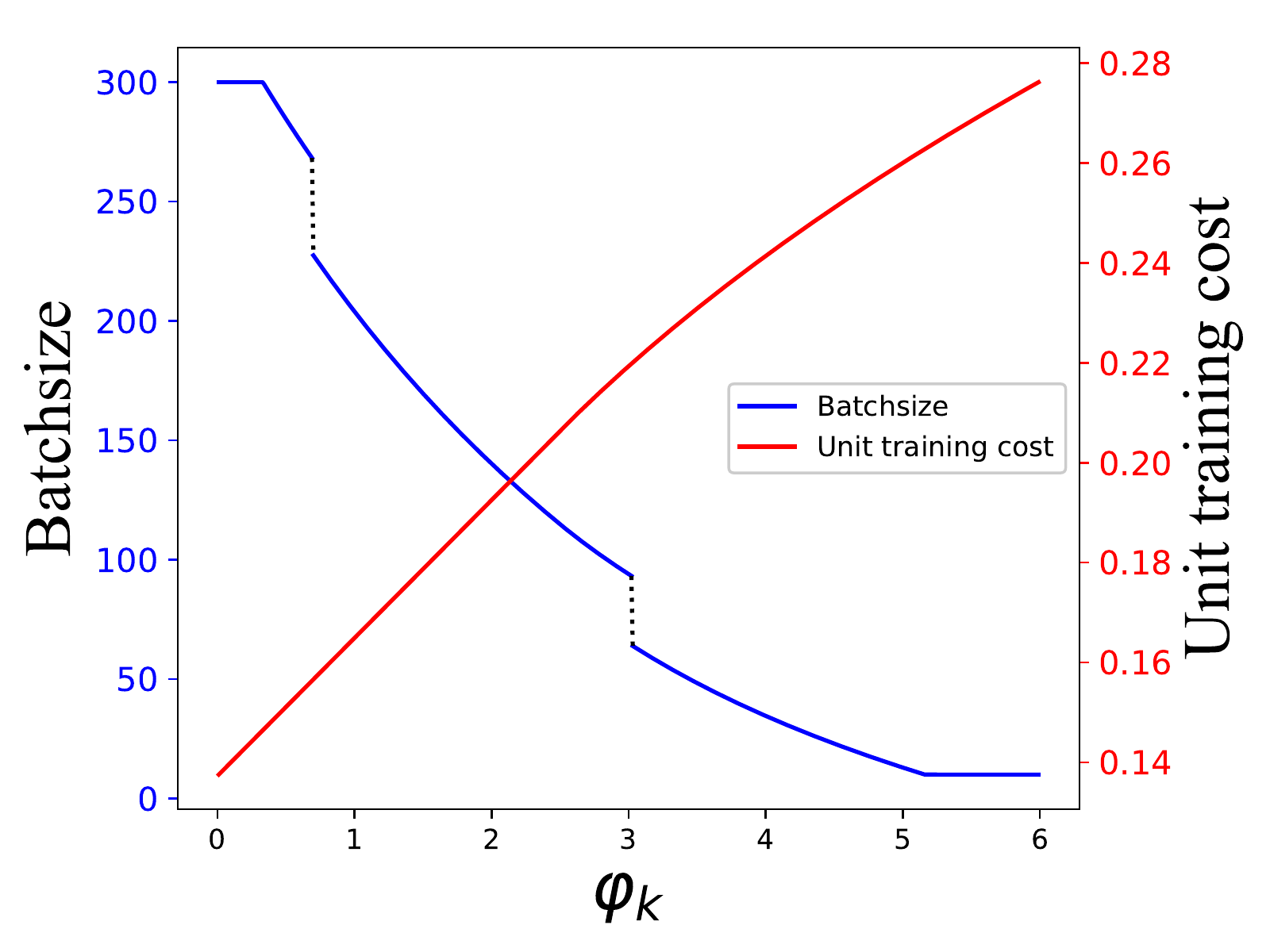}
		\caption{Impact of $\varphi_{k}$ on $B_{k}$}
		\label{batch_faic}
	\end{minipage}
	\begin{minipage}[t]{0.24\textwidth}
		\setlength{\abovecaptionskip}{-0.05cm}
		\centering
		\includegraphics[width=\linewidth]{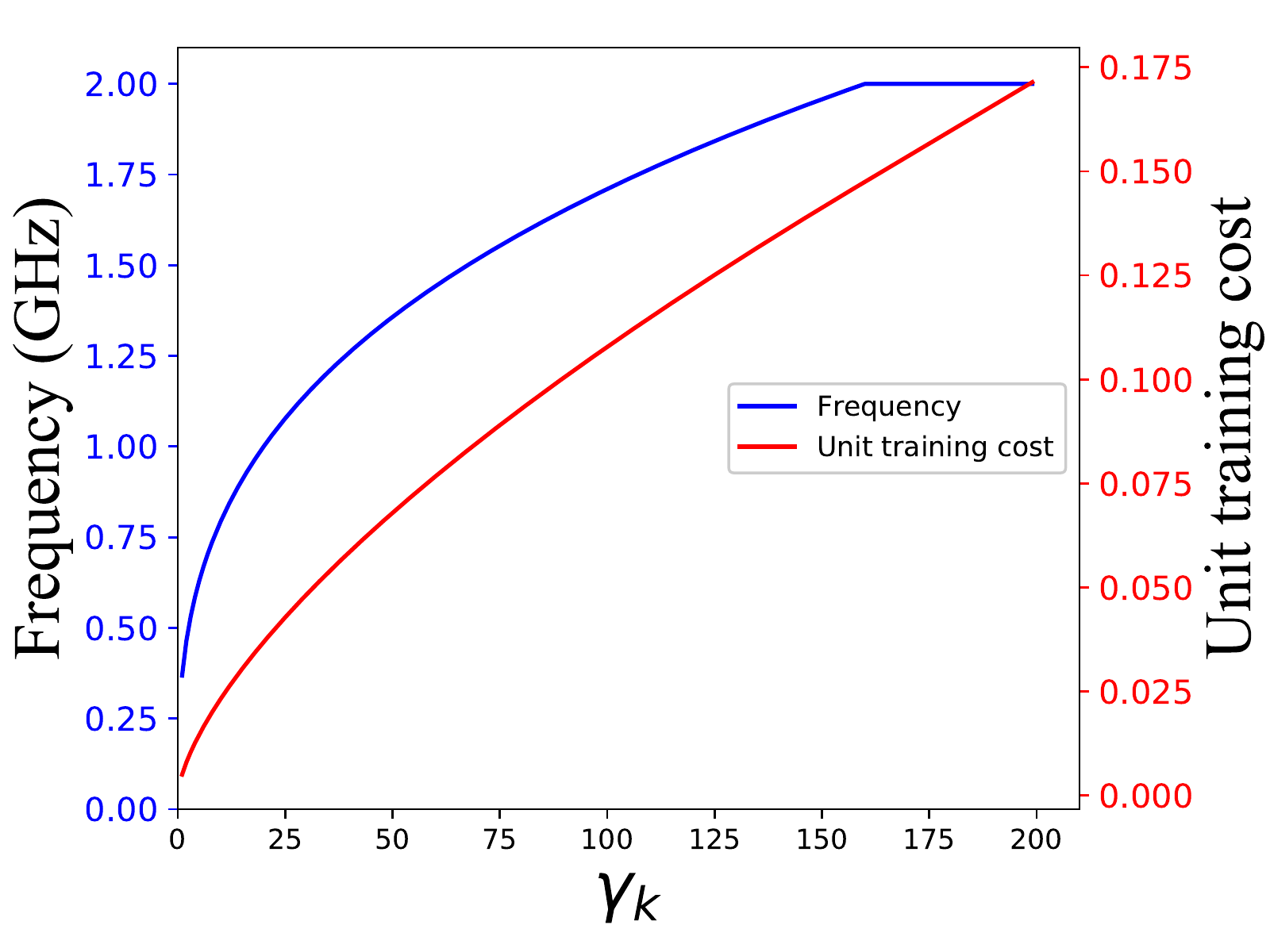}
		\caption{Impact of $\gamma_{k}$ on $A_{k}$}
		\label{batch_omegafre}
	\end{minipage}
\end{figure*}

\subsubsection{The performance of CAFL vs others schemes}
We compare with different schemes to study the superiority of the performance of CAFL. The details of each scheme can be listed as follows:
\textit{Uniform contribution}: Each of participant contributes the same amount of batchsize to maximize the total utility. \textit{Optimal total utility}: Maximize the utilities of all the users regardless of individual rationality and free-riding. \textit{Independent training}: Each of participant trains local model independently.
For experiment settings, each of group consists of 10\% \textit{Hq} and 90\% \textit{Lq} participants and run 100 times under each type of group. We average performance on global batchsize, total utility and contributing participants relative to the number of participants for each scheme (Fig. \ref{compare_batch}, \ref{compare_profit}, \ref{compare_participant}).

In Fig.\ref{compare_batch} and \ref{compare_profit}, the CAFL is superior of \textit{Uniform contribution} and \textit{Independent training} both in the global batchsize and total utility. Even though \textit{Uniform contribution} can alleviate free-riding problem, it inevitably stifles the high quality participants' motivation on training.
In terms of total utility (Fig. \ref{compare_profit}), the performance of CAFL is within 92\% of \textit{Optimal total utility}. As number of participants increase, the difference between \textit{Uniform contribution} and \textit{Optimal total utility} becomes larger and larger, while the difference between CAFL and \textit{Optimal total utility} is stable. Obviously, CAFL can be more efficient in large-scale FL application.

In Fig. \ref{compare_participant}, the contributing participants of \textit{Optimal total utility} is the lowest. This is because it selects a part of high quality participants to serve all participants to achieve the highest total utility. \textit{Independent training} is the highest one (as highest baseline), since each participant has motivation on training the local model ($\beta_{k}>0$). By observing Fig. \ref{compare_participant}, the performance of CAFL is slightly lower than that of \textit{Uniform contribution}, since CAFL removes a part of the lower quality participant to hold a result with higher total utility. It can be seen that CAFL is a good trade-off between total utility and contributing participants.

\subsection{Participants' Behaviors}
In this subsection, we study the impact of parameters on participants' strategies and analyze participants' behavior in CAFL.

For the intuitiveness of the experimental results, we use a group, which includes 10 \textit{High quality} and 10 \textit{Low quality} participants. Hence, all participants' $B_{k}^{max}$ in group range to $[30,150]$.
In the following experiments, we modify the parameters of the $20^{th}$ participant (current participant) with the lowest $\beta_{k}$ in the group.

We first consider a CAFL with $B_{th}=10$. Current participant's $\theta_{k}$ is uniformly distributed in $[80, 160]$. We set $\varphi_{k}=10$ and $\gamma=50$. The maximum batchsize of the current participant $B_{k}^{max}$ is 300. Other parameters of the participant are the same as those in the group. Fig. \ref{batch_theta} shows the relation between the strategy and $\theta_{k}$. Intuitively, a larger $\theta_{k}$ means higher preference about FL model, which indicates participants are willing to participate in the training with a large batchsize.
We observe that the value of batchsize has some big jumps (dotted line) occasionally with the increase of $\theta_{k}$, since relative sizes of current participants' $\beta_{k}$ and others' have changed, which reflects participant's position in a community influences his strategy dramatically.
Due to the characteristic of public goods, participants with higher preference about model will try their best to contribute more training data, which means these lower preference participants' needs are satisfied to some extend. Thus, these lower preference participants significantly decrease their contributions, which urges the current participant to dramatically increase the training data to meet his high demand.

In Fig. \ref{batch_faif} and Fig. \ref{batch_faic}, $\varphi_{k}$ ranges from 0.01 to 6. The parameter $\theta_{k}$ equals to 150. $\gamma_{k}=167.56$ is identical to that in the group. $B_{th}$ and $B_{k}^{max}$ are 10 and 300 respectively. In Fig. \ref{batch_faif}, as $\varphi_{k}$ increases, the participant is more concerned about energy consumption and attempts to decrease CPU frequency to minimize unit training cost. Fig. \ref{batch_faic} shows that the batchsize does not decrease linearly, since participant adjusts frequency to slow down unit training cost growth. Besides, $B_{k}$ plummets occasionally with the increase of unit training cost because of the change of the relative relationship in group.

We set $\varphi_{k}=1$ to show the impact of $\gamma_{k}$. $\gamma_{k}$ is uniformly distributed in $[1, 200]$. Here, $f_{k}^{max}=2$ GHz.
Fig. \ref{batch_omegafre} illustrates that the change rates of frequency and unit training cost respect to $\gamma_{k}$ are both numerically lower than those respect to $\varphi_{k}$. In reality, participants are more sensitive to energy consumption than training latency.

\section{Related work}
Google proposes federated learning framework using the federated average algorithm \cite{Com16}. This algorithm aims to aggregate model parameters or gradients from mobile devices without revealing their raw date. Summarizing concept and applications in federated learning, Yang \textit{et al.} classify federated learning into three types: horizontal federated learning, vertical federated learning and federated transfer learning \cite{yq2019}.

\textbf{FL performance and resourse optimization}:
Tran \textit{et al.} consider the trade-off among model performance, latency and energy consumption in FL \cite{Fed19}.
To accelerate model training, Ren \textit{et al.} optimize allocation of communication resources and selection of local batchsize during training model \cite{Acc19}.
Li \textit{et al.} discuss the convergence of FL algorithm on non-iid data \cite{noniid19}. Khan \textit{et al.} first propose self-organized FL and discuss a optimization problem of global federated learning time without a centralized server \cite{self2020}.

\textbf{Incentive Mechanism in FL:} Most of studies are based on an assumption that all participants participate in FL unconditionally.
To attract more participants in federated learning, a well-designed incentive mechanism is necessary.
Kang \textit{et al.} apply contract theory to design an incentive mechanism to attract participants with high-quality data, i.e., high-quality data owners can receive more rewards \cite{contract2019}.
Kang \textit{et al.} combine reputation and contract theory to design a novel incentive mechanism to ensure reliable FL \cite{joint2019}.
Zhan \textit{et al.} design a deep reinforcement learning-based incentive mechanism to obtain the optimal strategies of central server and participating edge nodes \cite{DRL2020}.
Sarikaya \textit{et al.} considers the trade-off between training latency and payment for workers from the perspective of central server using stackelberg game \cite{Motivate2019}.
Pandey \textit{et al.} proposes a novel crowdsourcing
framework to attract participant clients to provide a local model with a certain accuracy \cite{Crow2019}.
Existing papers are mainly concerned about a centralized task publisher scenario where there is only one model owner attracting others to complete FL. Different from previous studies, we discuss a self-organized FL where independent participants organize a community to collaborate on building a shared model in order to ensure a stable and fair federated learning system.

\textbf{Game theory}: Game theory is a powerful tool to analyze the situation where many participants make optimal decisions considering effect from others' strategies. It has been successfully applied in data acquisition \cite{Incent12}, data privacy preservation \cite{Game2015}, incentive mechanism and resources optimization in FL \cite{Motivate2019} \cite{Re2019}. There are some common approaches to achieve a Nash equilibrium such as decision trees \cite{Game2015}, best response dynamics, solving decision makers' best-response functions simultaneously \cite{Gamebook}.

\section{Conclusion}
In this paper, we develop a comprehensive theoretical framework for analyzing participants' behaviors in participant-centric federated learning. We propose the COFL game model and achieve the Nash equilibrium. To alleviate the free-riding phenomenon in COFL, we propose CAFL game model and establish a minimum threshold mechanism, which achieves the desired advantages of incentive-driven collaboration, free-riding mitigation and global efficiency boosting for participant-centric FL. We further show that optimal contribution threshold based CAFL game solution can significantly boot the amount of participation and system performance.

For the future work, we are going to study the implementation issue with truthful information reporting by globally fine-tuning the contribution threshold to balance the truthfulness and optimality. We will further address the issues of model security and non-IID data for participant-centric FL.


%





%
\bibliographystyle{IEEEtran}
\bibliography{myreference}

%








\end{document}